\documentclass[10pt]{article}
\usepackage{fullpage}
\usepackage{times,amsmath,amsfonts,amssymb,url,color,graphicx,
  mathtools}
\usepackage{natbib}

\usepackage{tikz}
\usepackage{ifthen}   
\usetikzlibrary{trees, positioning}

\newcommand{\reals}{\mathbb{R}}
\DeclareMathOperator*{\E}{\mathbb{E}}
\DeclareMathOperator*{\prob}{\mathbb{P}}

\newcommand{\inner}[1]{\langle #1 \rangle}

\newtheorem{theorem}{Theorem}
\newtheorem{definition}{Definition}

\newtheorem{lemma}{Lemma}
\newtheorem{corollary}{Corollary}
\newtheorem{remark}{Remark}
\newcommand{\BlackBox}{\rule{1.5ex}{1.5ex}}  
\newenvironment{proof}{\par\noindent{\bf Proof\ }}{\hfill\BlackBox\\[2mm]}

\newtheorem{example}{Example}

\newcommand{\lemref}[1]{Lemma~\ref{#1}}
\newcommand{\thmref}[1]{Theorem~\ref{#1}}

\newcommand{\secref}[1]{Section~\ref{#1}}
\newcommand{\figref}[1]{Figure~\ref{#1}}
\renewcommand{\eqref}[1]{Equation~\ref{#1}}

\newcommand{\Var}{\operatorname{Var}}

\newcommand{\Unif}{\operatorname{Unif}}
\newcommand{\ip}[1]{\left\langle #1 \right\rangle}

\title{From Reasoning to Super-Intelligence: A Search-Theoretic Perspective}
\author{Shai Shalev-Shwartz and Amnon Shashua}
\date{AA-I, July 2025}

\begin{document}

\maketitle

\begin{abstract}
  Chain-of-Thought (CoT) reasoning has emerged as a powerful tool for
  enhancing the problem-solving capabilities of large language models
  (LLMs). However, the theoretical foundations of learning from CoT
  data remain underdeveloped, and existing approaches—such as
  Supervised Fine-Tuning (SFT), Reinforcement Learning (RL),
  Tree-of-Thoughts (ToT), and Monte Carlo Tree Search (MCTS)—often
  fail on complex reasoning tasks. In this work, we identify core
  obstacles that hinder effective CoT learning, including distribution
  drift, lack of embedded search, and exponential inference costs. We
  introduce the Diligent Learner, a new learning paradigm that
  explicitly models reasoning as a depth-first search guided by a
  validator and supports backtracking upon failure. Under two mild and
  realistic assumptions, we prove that the Diligent Learner can
  efficiently learn from CoT data while existing methods fail to do so. This
  framework offers a path toward building scalable and reliable
  reasoning systems trained on naturally occurring, incomplete
  data—paving the way for the development of Large Reasoning Models
  (LRMs) with robust, interpretable problem-solving abilities.
\end{abstract}

\section{Introduction}

The capacity for reasoning lies at the core of the pursuit of intelligence. Although contemporary Large Language Models (LLMs) and Large Reasoning Models (LRMs) exhibit non-trivial reasoning capabilities, even the most advanced models continue to fall short in practice. Specifically, as the difficulty of a task surpasses a certain threshold, the performance of these models degrades sharply, often resulting in a complete failure to solve the problem (\cite{stechly2025,shojaee2025,hochlehnert2025}).

We begin with the observation that complex reasoning tasks frequently require a search process over a space of abstract ideas. In such scenarios, it is seldom the case that a reasoning step follows deterministically from previous steps. Instead, the reasoner is typically confronted with uncertainty and must engage in trial-and-error exploration across a subset of promising directions. This uncertainty is generally resolved only in hindsight, once a particular exploratory trajectory is either validated or refuted. In such cases, a backtracking operation to a prior point of deliberation becomes necessary. Expert human reasoners often maintain a structured mental representation of this process, akin to a reasoning search tree, while avoiding combinatorial explosion through selective and efficient exploration.

As an illustrative example, consider the task of proving that no coprime positive integers \( a, b, c \) satisfy the equation \( a^2 + b^2 = 6c^2 \). An LLM trained on a comprehensive corpus of number theory texts and problems may propose useful intermediate steps. One such step could involve the use of modular arithmetic to restrict the solution space to a finite number of cases in order to derive a contradiction. A domain expert would likely select a modulus \( k \) informed by the structure of the equation, such as a prime divisor of the coefficient 6, namely \( k = 2 \) and \( k = 3 \). Figure~\ref{fig:coprime} illustrates a possible reasoning process that involves a sequence of explorations and refinements, culminating in a successful proof. This process can be naturally modeled as a tree whose root represents the initial problem statement and whose leaf nodes represent terminal outcomes. In this specific example, the successful reasoning path corresponds to the sequence of nodes \( \{0,6,9,10,11,12\} \), terminating in a node labeled ``Done''.

This example highlights several important features of the reasoning process. Each path from the root to a leaf corresponds either to a valid chain of reasoning leading to a correct solution—hereafter referred to as a ``golden path''—or to an unsuccessful exploration. Importantly, failed paths can typically be ruled out only in retrospect, after reaching a terminal point along the path. For instance, when situated at node 0, one cannot eliminate the subpath beginning with node 1 without further investigation. Similarly, from the vantage point of node 6, the viability of the path starting at node 7 remains indeterminate until additional steps are considered. This inherent non-determinism is a defining characteristic of reasoning tasks beyond a certain threshold of complexity.

The second important lesson we can learn from the example in
Figure~\ref{fig:coprime} concerns the nature of available
supervision. While it is plausible to find this problem—along with a
clean, step-by-step golden path to its solution—in a textbook or
online resource, it is exceedingly rare to find a record of the full
reasoning process that includes false starts, failed attempts, and
backtracking. The final successful trajectory is typically all that is
preserved, while the underlying search that led to it is discarded or
never externalized. This asymmetry reflects a broader reality:
human-authored reasoning data overwhelmingly consists of polished,
linear solutions, with little to no evidence of the underlying
exploration.

This presents a key challenge for learning. Because reasoning tasks
inherently involve uncertainty and delayed feedback—where the validity
of a path is often clear only in hindsight—learning solely from golden
paths is insufficient. To acquire robust reasoning capabilities, a
learner must develop mechanisms for self-guided exploration:
generating and evaluating its own reasoning attempts, learning from
failure, and adjusting course when needed. Building such reflective
capabilities is essential for navigating the complexities of
real-world reasoning.

The terms ``Chain-of-Thought'' (CoT)~\citep{wei2022chain} or ``Scratchpad''~\citep{nye2021show} are commonly used to describe successful sequences of reasoning steps. Given that golden-path CoTs constitute the primary form of available supervision, the question of CoT learnability—namely, identifying the properties of CoTs that make them amenable to learning—emerges as a central issue. This naturally leads to the inquiry: Can existing learning algorithms effectively learn from CoTs, and if not, what alternative frameworks might be more suitable?

Although the learnability of CoTs has been examined in prior work~\citep{Malach2023, joshi2025theory}, those investigations are based on assumptions that diverge from the characteristics of real-world reasoning data. In particular, they often presuppose the existence of a single, deterministic hypothesis capable of predicting the exact next token in a reasoning sequence. This assumption is misaligned with the nature of human-generated data, which is inherently diverse. Semantically equivalent reasoning steps may be expressed through syntactically distinct formulations. Furthermore, as exemplified earlier, multiple plausible continuations may exist at each step in the reasoning process, and it is typically not known a priori which continuation leads to a correct solution. Thus, the presence of uncertainty and linguistic variability renders deterministic token-level prediction inadequate for capturing the richness of natural language reasoning.

The first contribution of the present work is to propose a set of
sufficient conditions for CoT learnability that are likely to hold for
naturally occurring data. It is then shown that existing learning
algorithms might provably fail even when these conditions hold. In
other words, these conditions are insufficient for existing
algorithms. Specifically, current methods suffer from one or more of
the following deficiencies: susceptibility to distribution drift, the
absence of an embedded search mechanism, or exponential complexity
during training or inference.

The second contribution is the introduction of a novel learning algorithm, referred to as the \textit{diligent learner}, designed for training LRMs under the constraints imposed by CoT supervision. This algorithm is provably effective under the proposed learnability conditions. It avoids the aforementioned limitations, incorporates a search mechanism as an intrinsic component, operates exclusively on golden-path CoTs, and demonstrates strong performance on reasoning tasks of significant difficulty.

Finally, by establishing a training methodology that supports CoT learnability, this work opens the possibility for artificial agents to construct and navigate reasoning trees of depth and breadth exceeding human cognitive capacities. In this sense, the diligent learner may represent a concrete step toward the realization of super-intelligent systems.

\begin{figure}
\begin{tikzpicture}
  [
    grow=down,
    sibling distance=12cm,
    level distance=4cm,
    edge from parent/.style={draw, -latex},
    treenode/.style = {
      rectangle, rounded corners, draw,
      align=center, minimum height=3cm,
      fill=blue!10, text width=5cm,
      font=\small
    },
    idstyle/.style = {
      circle, draw, fill=white,
      inner sep=0pt,
      minimum size=5mm,
      font=\scriptsize
    }
  ]

  \node[treenode] (n0) {Prove: there are \textbf{no} coprime positive integers
    $a,b,c$ s.t. $a^2+b^2=6c^2$.}
    child { node[treenode] (n1) {Let's show that strict inequality must hold.}
      child { node[treenode] (n2) {All 3 numbers are coprime, hence
          they must be different. Assume w.l.o.g. $a < b$, and
          consider the 3 cases for which $c > b$, $c \in (a,b)$, $c
          < a$} 
        child { node[treenode] (n3) {Case 1: $a^2+b^2 < 2c^2 <
            6c^2$, we get strict inequality as desired.}
          child { node[treenode] (n4) {Case 3: $c < a$ but it tells
              us nothing about the relation between $a^2+b^2$ and
              $6c^2$, so we are stuck. }
            child { node[treenode] (n5) {Backtrack to node (0)}}
          }
        }
      }
    }
    child [sibling distance=6cm] { node[treenode] (n6) {Show that for some $k$, LHS mod k doesn't equal to
        RHS mod $k$, hence equality can't hold.}
      child { node[treenode] (n7) { Pick $k=2$. RHS is even. So LHS
          must be even as well. The parity of $a^2$ is the same as
          $a$. Since $a,b$ are coprime, one of them
          must be odd. But maybe both are odd, so there's no
          contradiction.}
        child{ node[treenode] (n8) {Backtrack to (6)} }
      }
      child { node[treenode] (n9) { Pick $k=3$. RHS is divisible by
          $3$ hence RHS mod 3 = 0. }
        child { node[treenode] (n10) { For every $a$, if $a~
            \mathrm{mod}~3=0$ then so is $a^2$, and otherwise, write it as
            $a=3n+r$ for $r \in \{1,2\}$, then $a^2 ~\mathrm{mod}~ 3 =
            r^2 ~\mathrm{mod} ~3 =1$. }
          child { node[treenode] (n11) {Since $a,b$ are coprime, at
              least one of them shouldn't be divisible by $3$. Hence
              $a^2+b^2~\mathrm{mod}~3 \in \{1,2\}$.}
            child { node[treenode] (n12) { We have shown
                that LHS mod 3 is in $\{1,2\}$ while RHS mod 3 is
                $0$. \textbf{Done.} }
            }
          }
        }
      }
    };

  \node[idstyle] at (n0.north west) {0};
  \node[idstyle] at (n1.north west) {1};
  \node[idstyle] at (n2.north west) {2};
  \node[idstyle] at (n3.north west) {3};
  \node[idstyle] at (n4.north west) {4};
  \node[idstyle] at (n5.north west) {5};
  \node[idstyle] at (n6.north west) {6};
  \node[idstyle] at (n7.north west) {7};
  \node[idstyle] at (n8.north west) {8};
  \node[idstyle] at (n9.north west) {9};
  \node[idstyle] at (n10.north west) {10};
  \node[idstyle] at (n11.north west) {11};
  \node[idstyle] at (n12.north west) {12};

\end{tikzpicture}
\label{fig:coprime}
\end{figure}

\section{CoT learnability: A model of Learning from Chain-of-Thought} \label{sec:main}

We consider learning problems in which the input space, $\mathcal{X}$,
and the output space, $\mathcal{Y}$, are both the set of sequences of
tokens of length at most $n$. That is,
$\mathcal{X} = \mathcal{Y} = \Sigma^{\le n}$, where $\Sigma$ is our
alphabet of tokens.  An instance of a learning problem is defined by a
distribution $D_x$ over $\mathcal{X}$ and a labeling function
$f : \mathcal{X} \to 2^{\mathcal{Y}}$. For every $x$, the subset
$f(x) \subset \mathcal{Y}$ is the set of text pieces that describe a
``correct'' response to the input $x$. This setting captures typical
problems in science and engineering. The size of $f(x)$ may be very
large. Think for example of $x$ as an algorithmic problem and of $y$
as a source code implementing a solution to the problem, then there
may be many different implementations of a correct code. We 
therefore do not assume that the learner receives the entire set
$f(x)$.

The learner has access to two types of learning feedback. In the first
one, the learner has an access to $m$ examples
$(x_1,y_1),\ldots,(x_m,y_m)$ where each $x_i$ is sampled i.i.d. from
$D_x$ and each $y_i$ is some element from $f(x)$. Learning based on
this feedback is often called Supervised Learning. Furthermore, when
one performs Stochastic Gradient Descent (SGD) iterations on the $m$
examples while initializing the model by a pre-trained foundation
model, then the learning process is often called Supervised
Fine-Tuning (SFT).  In the second model, the learner receives
$x_1,\ldots,x_m$, again sampled i.i.d. from $D_x$, and has an oracle
access to a reward function such that for every $i$ and
$y \in \mathcal{Y}$, the reward returns $r(x_i,y) = 1[y \in
f(x_i)]$. That is, the reward returns 1 if $y$ is a correct response
to $x_i$ and $0$ otherwise. Learning with such a reward function is
often called Reinforcement Learning with Verified Rewards (RLVR).

We focus on learning with auto-regressive \textbf{generative models}, such as
Large-Language-Models (LLMs) and Large-Reasoning-Models (LRMs).
An auto-regressive generative model is based on a function,
$\pi_\theta(t_i|t_0,\ldots,t_{i-1})$, that defines the probability to
generate the next token given previous ones. For concreteness, one can
think of the function $\pi_\theta$ as being a transformer neural
network with $\theta$ being a vector representing the weights of the
network. However, most of our results can be generalized to an
arbitrary form of $\pi_\theta$. Applying $\pi_\theta$
auto-regressively, token-by-token, with a maximum of $n$ generations,
induces a conditional probability over $\mathcal{Y}$ for every given
input $x$. It follows that $\pi_\theta$ induces a function
$h_\theta : \mathcal{X} \to \Delta(\mathcal{Y})$, where
$\Delta(\mathcal{Y})$ is the set of probabilities over $\mathcal{Y}$.

Since $h_\theta(x)$ is a probability over $\mathcal{Y}$, defining its
``correctness'' depends on an acceptable threshold for generating a
correct answer out of it. Indeed, when practitioners evaluate
generative models, they sometimes report the pass\texttt{@}k metric,
aiming at approximating the likelihood that the generative model would
generate a correct output within $k$ attempts. Formally, we say that
$h_\theta(x)$ is $\gamma$-correct if
\begin{equation} \label{eqn:gammaCorrect}
\prob_{y \sim h_\theta(x)}[y \in f(x)] \ge \gamma ~.
\end{equation}
Note that if $h_\theta(x)$ is $\gamma$-correct, then on average it
would succeed in $k = 1/\gamma$ generation attempts. 

Numerous success stories attest to the empirical effectiveness of SFT
(see for example \cite{bubeck2023sparks}).  However, negative
empirical evidence has also been quickly gathered, showing that this
approach fails for some natural problems, especially ones that when
solved by humans, require 'step-by-step' thinking.  As a simple and
natural example consider the problem of learning to multiply two
numbers. The input is comprised of two strings of digits $a,b$,
representing integers $N(a),N(b)$ in the usual decimal basis, and the
network should output a string of digits forming the decimal
representation of $N(a) \cdot N(b)$, one digit at a time.  While
humans learn to solve this problem in elementary school, we prove (in
section~\ref{sec:multiplication}) that using SGD over an
auto-regressive transformer architecture for this task would require
an exponentially large number of gradient steps (in the number of
digits). The proof, which may be interesting by itself, relies
on a generalization of Parseval inequality for functions with multiple
outputs that are $\epsilon$-approximately-orthonormal.  We represent
functions using Fourier characters, allowing us to reduce inner
products to scalar characters, and to use the permutation-invariant
nature of transformer networks to bound the variance of a family of
``multiplication-like'' problems.

As mentioned previously, humans learn multiplication at elementary
school. However, they do not learn this task in an end-to-end
manner. Instead, they are taught to solve multiplication by following
the long multiplication algorithm step-by-step. \cite{wei2022chain}
has shown that prompting LLMs to think ``step-by-step'' significantly
improves their ability to perform complex reasoning. This
``step-by-step'' thinking is also called Chain-of-Thoughts (CoT).
Is multiplication efficiently learnable with CoT supervision?
\cite{liu2023goat} have shown
empirically that while SGD over a transformer architecture fails to
learn multi-digit multiplication in an end-to-end manner, it succeeds
to learn it when providing CoTs that decompose the multiplication task
into a series of easier tasks by leveraging basic arithmetic
principles. 

This paper tackles two fundamental questions regarding learning with CoTs:
\begin{enumerate}
\item \textbf{What is efficiently learnable with CoTs?} In particular,
  what properties of CoTs make a problem efficiently learnable? And,
  do these properties hold for natural large scale data?
\item \textbf{How to learn with CoTs?} In particular, can existing
  learning algorithms learn with CoTs? And if not, is there an
  alternative algorithm that can learn with CoTs?
\end{enumerate}

In CoT learning, the learner receives $m$ examples, but now each
example is a triplet $(x_i,z_i,y_i)$, where
$x_i,y_i \in \Sigma^{\le n}$ are the texts describing the input and
output, and $z_i \in \Sigma^*$ is a text describing the CoTs.

In our study of CoT learning, we make a fundamental distinction
between \emph{validation} and \emph{search}. The \textbf{validation
  problem} is defined as follows: given a triplet $(x, z, y)$, the
goal is to determine whether the reasoning chain $z$ successfully
leads from input $x$ to a correct output $y$. This can be formalized
as a binary classification task that returns $1$ if $(z, y)$ is a
valid explanation and a correct answer for the input $x$, and $0$
otherwise. In contrast, the \textbf{search problem} takes as input
only $x$ and asks the model to generate a valid reasoning-output pair
$(z, y)$ such that the triplet $(x, z, y)$ passes the validation check.

A key observation, rooted in computational complexity theory, is that
the validation problem is often substantially easier than the search
problem. This asymmetry lies at the heart of the complexity class NP
(nondeterministic polynomial time): a problem is in NP if, given a
proposed solution, one can efficiently verify its correctness, even if
finding such a solution from scratch may be computationally hard. For
instance, consider a complex math word problem. Once a student
provides a detailed step-by-step derivation and an answer, a teacher
(which will be our automated validation system) can quickly verify
whether each step is logically sound and the final answer is
correct. However, producing such a derivation from scratch -- the
search task -- may require significant time, creativity, or insight.

Motivated by this distinction, we assume an access to a validator -- a
mechanism that can reliably assess whether a given $(x, z, y)$ is
valid. The validator can itself be learnt from examples.  A detailed
approach as to how to validate reasoning is outlined in our previous
paper on PAC reasoning~\citep{shalev2024artificial}. This allows us to
shift focus away from validation and toward the search problem, which
is central to training generative models for problem
solving. Specifically, our goal is to design and analyze learning
algorithms that can efficiently generate correct and interpretable
reasoning chains and solutions, leveraging the validator as a tool to
guide and supervise the search process.

At a high level, regarding the ``what'' question, we argue that
existing theory of CoT learning imposes strong assumptions that are
unlikely to hold for natural data. Instead, we provide less
restrictive sufficient properties and argue that they are very likely
to hold for natural data. Regarding the ``how'' question, we first
prove that existing learning algorithms might fail to learn certain
problems, even problems that can be learnt with alternative means. We
then present our \emph{diligent learner}, a new general algorithm for
CoT learning.

\section{What is efficiently CoT-learnable?}
Existing theory by \cite{Malach2023, joshi2025theory} shows
succesful CoT learning under the following assumption: consider a
hypothesis class, $\mathcal{H}$, such that each $h \in \mathcal{H}$ is
a mapping from $\Sigma^*$ to $\Sigma$ and assume that there exists
$h^* \in \mathcal{H}$ such that for every example $i$, and every token
$t_{i,j}$ in $(z_i,y_i)$, we have that
$t_{i,j} = h^*(x,t_{i,1},\ldots,t_{i,j-1})$. Then, if there is an
efficient PAC learner for $\mathcal{H}$, then by invoking this learner
on all of the tokens in all of the CoT examples, we would get a
hypothesis $h \in \mathcal{H}$ that when applied auto-regressively
would generate a succesful CoT.

This result can explain the success of learning multiplication when
the CoT is a trace of the long multiplication algorithm. In fact,
\cite{Malach2023} has shown that the above assumption holds whenever
$z_i$ is the trace of any efficient \textbf{deterministic} algorithm
that maps from $x_i$ to $y_i$ (and long multiplication is an example
of such algorithm). But in most cases, having a trace of an efficient
deterministic algorithm that maps from $x_i$ to $y_i$ means that we
already have that algorithm so why bother learning it.

We argue that the assumption that there exists a single deterministic
hypothesis that can always predict the exact next token in a reasoning
sequence is incompatible with the nature of real-world data,
particularly the kind of data used to train LLMs. Human-generated data
is intrinsically diverse in phrasing, structure, and granularity. For
instance, one step in a proof may be stated in many logically
equivalent but syntactically distinct ways--e.g., ``apply mod 3 and
reach contradiction'' vs. ``reduce both sides modulo 3; contradiction
follows.'' Furthermore, as illustrated in our running example in
\figref{fig:coprime}, and as argued before, many steps in the
reasoning chain involve \textbf{search and uncertainty}—the learner
may not know which of multiple plausible continuations leads to a
solution until exploring further. As a result, there is inherent
uncertainty in the continuation of a reasoning chain, and demanding
exact token-level prediction ignores the variability and richness in
natural language reasoning.

We see that the existing CoT theory does not capture the true nature
of reasoning. To develop a different theory, we first
shift the unit of prediction from individual tokens to semantic steps
in a reasoning chain. That is, rather than requiring the learner to
predict the next token $t_{i+1}$ given the prefix $t_1,\ldots,t_i$, we
instead model the reasoning trace as a sequence of coherent steps:
$x = v_1, v_2, \ldots, v_T = y$, where each $v_i \in \Sigma^{\le n}$
corresponds to a meaningful intermediate assertion or derivation. This
coarse-grained view captures the actual structure of reasoning better
than fine-grained token sequences.

To state our sufficient conditions for CoT learnability, let us first
recall the definition of a Probably-Approximately-Correct (PAC)
learner, with slight adaptions to our setting.
\begin{definition}[PAC learner]
  Let $\mathcal{I}, \mathcal{O} \subset \Sigma^{*}$ be an input and
  output domains and 
  let $\mathcal{F}$ be a set of functions from $\mathcal{I}$ to
  $\mathcal{O}$.
  We say that
  $\mathcal{A}$ is an efficient PAC learner for $\mathcal{F}$ if for
  every $\epsilon,\delta \in (0,1)$ and $n \in \mathbb{N}$, there
  exists $m = \mathrm{poly}(n,1/\epsilon,\log(1/\delta))$ such that
  the following holds for every $f \in \mathcal{F}$ and every distribution
  $\mathcal{D}$, where both the support of $\mathcal{D}$ and the range
  of $f$ is contained in $\Sigma^{\le
    n}$
  \begin{itemize}
  \item The algorithm $\mathcal{A}$ runs in time $O(m)$ and uses at most $m$
    examples
  \item Each example $(x_i,y_i)$ is generated by sampling i.i.d. $x_i
    \sim \mathcal{D}$ and setting $y_i = f(x_i)$
  \item With probability of at least $1-\delta$ over the sampling of
    the $x_i$'s, $\mathcal{A}$ returns a hypothesis $h : \mathcal{I}
    \to \mathcal{O}$ such that
    \[
      \prob_{x \sim \mathcal{D}}[h(x) = f(x)] \ge 1-\epsilon
    \]
  \end{itemize}
\end{definition}

We extend the above definition of PAC learner to deal with learning of
generative models, while relying on the notion of $\gamma$-correctness
as given in \eqref{eqn:gammaCorrect}. There are several major
differences between the definitions. First, the range of the functions in
$\mathcal{F}$ is now $2^{\mathcal{O}}$ instead of $\mathcal{O}$. This
allows multiple correct answers to every input. Secondly, the range of
the learnt
functions is now $\Delta(\mathcal{O})$ instead of $\mathcal{O}$,
capturing the fact that we learn a generative model rather than a
deterministic function. Finally, the notion of correctness is now
$\gamma$-correctness as opposed to absolute correctness. 
\begin{definition}[Generatively-Probably-Approximately-Correct (GPAC) learner]
  Let $\mathcal{I}, \mathcal{O} \subset \Sigma^{*}$ be an input and
  output domains and let $\mathcal{F}$ be a set of functions from
  $\mathcal{I}$ to $2^{\mathcal{O}}$.  For a fixed $\gamma \in (0,1]$,
  we say that $\mathcal{A}$ is an efficient $\gamma$-GPAC learner for
  $\mathcal{F}$ if for every $\epsilon,\delta \in (0,1)$ and
  $n \in \mathbb{N}$, there exists
  $m = \mathrm{poly}(n,1/\epsilon,\log(1/\delta))$ such that the
  following holds for every $f \in \mathcal{F}$ and every distribution
  $\mathcal{D}$ over $\mathcal{I} \times \mathcal{O}$, s.t. the
  marginals of $\mathcal{D}$ w.r.t. $\mathcal{I}$ and $\mathcal{O}$
  are contained in $\Sigma^{\le n}$ and the conditional
  $\mathcal{D}(o|x)$ is non zero iff $o \in f(x)$
  \begin{itemize}
  \item The algorithm $\mathcal{A}$ runs in time $O(m)$ and uses at most $m$
    examples
  \item Each example $(x_i,y_i)$ is generated by sampling i.i.d. from
    $\mathcal{D}$
  \item With probability of at least $1-\delta$ over the sampling of
    the examples, $\mathcal{A}$ returns a hypothesis $h : \mathcal{I}
    \to \mathcal{O}$ such that
    \[
      \prob_{x \sim \mathcal{D}}\left[ \prob_{o \sim h(x)}[h(x) \in
        f(x)] \ge \gamma \right] \ge 1-\epsilon
      \]
  \end{itemize}
\end{definition}

We are now ready to state our sufficient conditions for CoT
learnability. We would \emph{not} assume that we can (efficiently) PAC
learn how to generate $v_i$ given $v_1,\ldots,v_{i-1}$ in a unique,
deterministic, manner, as this is not a realistic assumption. Instead,
our first assumption is that we can (efficiently) $\gamma$-GPAC learn how to
generate $v_i$ with a fixed probability parameter $\gamma$, which may
be strictly smaller than $1$. Intuitively, while it is unlikely that
we can learn to generate a unique correct $v_i$, it is very likely
that we can learn a generative model, whose likelihood to generate
some correct $v_i$ (among many possibilities) is at least $\gamma$.

Due to this relaxed functions, we are going to have many generations
of incorrect chains.  We say that a reasoning chain $v_1,\ldots,v_T$
is \emph{correct and complete} if our validator accepts
$v_2,\ldots,v_T$ as a valid explanation and a correct answer for input
$x$. We say that $v_1,\ldots,v_t$ is \emph{correct} (but incomplete),
if it can be completed to a correct and complete reasoning chain. A
chain is incorrect if it cannot be completed to a correct and complete
reasoning chain of length at most $T_{\max}$ (where $T_{\max}$ is a
domain dependent parameter). Finally, for an incorrect chain,
$v_1,\ldots,v_t$, we denote by $\beta(v_1,\ldots,v_t)$ the maximal
index $i$ such that $v_1,\ldots,v_i$ is correct.

In light of the above, our second assumption is that when an incorrect
chain becomes long enough, we can detect that we are on the wrong
track and efficiently PAC learn to \textbf{backtrack} into the correct
prefix of the chain. Following the distinction between ``validation''
and ``search'', here again, while it may be difficult to know what is
the right next step as a foresight, in hindsight, it is much easier to figure
out that we are on the wrong track and how to get back to sanity.
All in all, this leads us to the following two milder sufficient
conditions for CoT learnability:
\begin{theorem}  \label{thm:main}
  The following two conditions are sufficient for CoT learnability.
\begin{itemize}
\item There is a fixed $\gamma \in (0,1]$, and an efficient
  $\gamma$-GPAC learner, such that for every $i$, it
  learns a generative
  model, that receives $v_1,\ldots,v_{i-1}$ and generates $v_i$.
\item There is an efficient PAC learner~\footnote{One can further
    relax this assumption, and requires only a sufficiently small error
    to ``overshoot'' (namely, to output a too small backtracking
    index), while we can tolerate ``undershooting'' that happens with a
    fixed probability over the incorrect chain.} that upon receiving
  as input a sufficiently long incorrect chain
  $v_1,\ldots,v_{T_{\max}}$ (or some incorrect prefix of it) learns to
  backtrack to $\beta(v_1,\ldots,v_{T_{\max}})$.
\end{itemize}
\end{theorem}

The proof of the theorem is constructive--- we present the
\emph{diligent learner} and prove that it manages to learn from CoT
data under the assumptions given in \thmref{thm:main}. This will be
presented in \secref{sec:diligent}. But before that, in the next
section, we overview existing algorithms and provably demonstrate
their limitations. 

\section{How \emph{not} to learn from CoTs} \label{sec:hownot}

We have described two properties of CoT data and claimed that they are
sufficient for learnability. This will be constructively proven in the
next section in which we describe and analyze our \emph{diligent
  learner}. Therefore, our diligent learner is one way to provably
learn with CoTs. Our goal in this section is to prove that existing
learning algorithms, including ones that were dedicatedly
designed for reasoning and search, might fail. Understanding the
failures of existing algorithms leads to the design principles of our
diligent learner. 

We start with baseline algorithms, and the immediate one is 
SFT on the training examples. We show two sources of failures:
\emph{distribution drift} and \emph{lack of search}.

\paragraph{Distribution drift:} The
\emph{distribution drift} problem stems from the fact that the
distribution of examples the learner observes at inference time might
diverge from the one the learner observed during training time. In the
language of the literature of Reinforcement Learning, SFT is also
called \emph{imitation learning}, since the learner aims at mimicking
the behavior of the teacher via observing the teacher's
actions. Distribution drift is a well known limitation of imitation
learning, since the learning is done in an open loop, whereas inference
involves multiple steps and errors might aggregate and drift the
learner to uncharted territory.

To formally demonstrate the distribution drift problem, observe that
SFT training of an auto-regressive model is done by optimizing (via
SGD iterations) a log-loss objective defined by \emph{teacher
  forcing}. This means that the objective is the average of the
log-loss over every token in each example given the previous tokens of
the same example. The term \emph{teacher forcing} indicates that while
at inference time, the model predicts the next token given its own
predictions of previous tokens in previous steps, during training, we
ignore the model past predictions and train it to predict the next
token given the previous tokens from the input ground truth data
(i.e. the teacher). In our context, since we do not assume that the
learner can perfectly and deterministically predict every next token
(instead, it can only be $\gamma$-correct), the examples seen at
inference time after a single error might look different than the
examples that were observed during training, leading to wrong future
predictions. We next build a synthetic example that demonstrates this
issue.

\begin{example} \label{example:drift}
Consider the following family of reasoning chains. The input, $x=v_1$,
is generated uniformly at random from $\{\pm 1\}^{2n}$. The next step
is $v_2 = \prod_{i=1}^{n} x_{\pi(i)}$ where $\pi$ is a fixed (but
unknown to the learner) permutation over $2n$. Next, for
$i \in \{3,\ldots,n-2\}$ we have that $v_i = v_{i-1}\,x_{\pi(i)}$. And
the output is $v_{n-2} = y$.
\end{example}

In the next section we prove that this problem is CoT learnable, and
in particular, our diligent learner learns to solve it
efficiently. Here we show that SFT would fail on this problem due to
distribution drift. Our analysis relies on known results for parity
learning with SGD (see \cite{wies2022sub} and the references
therein). Learning the step $v_2$ given $x$ by SGD is a parity with
$n$ out of $2n$ relevant bits problem, and it requires
$\Omega(\exp(n))$ SGD iterations. Using a smaller number of SGD
iterations would result in approximately 50\% accuracy when predicting
$v_2$. As to the rest of the chain, given that $v_{i-1}$ is correct,
predicting each $v_i$ involves just $2$ previous bits, and therefore
SGD will converge to the right rule very quickly. So, at training
time, SFT would succeed to (almost) perfectly learn all of the bits,
including the output bit, and would only struggle with learning
$v_2$. However, during inference time, an error in predicting $v_2$
(which happens with probability of roughly $1/2$) would contaminate
the rest of the chain, and would lead to a wrong output in roughly
$1/2$ of the time. This is a clear distribution drift problem since
the very same rule of predicting $y = v_{n-3} x_{\pi(n-2)}$ that works
perfectly on the training data (because the teacher forcing implies
that at training time $v_2$ is always correct) fails miserably on the
distribution over $v_2$ at inference time (because at inference time,
$v_2$ is only correct half of the times).

\begin{remark}
  Observe that for every $i > 2$,
  $v_i = v_2 \prod_{j=3}^i x_{\pi(j)}$. In particular, $y$ can be
  written as the parity of $4$ input bits as follows:
  $y = v_{n-2} = v_2 \prod_{j=3}^{n-2} x_{\pi(j)} = x_{\pi(1)}\,
  x_{\pi(2)} \, x_{\pi(n-1)}\, x_{\pi(n)}$. This representation of $y$
  doesn't suffer from distribution drift, since it only depends on the
  input. So, one may wonder why doesn't the SFT learner figure out
  that $y$ also equals 
  $x_{\pi(1)}\, x_{\pi(2)} \, x_{\pi(n-1)}\, x_{\pi(n)}$. The reason
  is a \emph{shortcut learning} problem---on the distribution of the
  training data, the rule $y = v_{n-3} x_{\pi(n-2)}$ is always correct,
  and is much simpler than the rule
  $y = x_{\pi(1)}\, x_{\pi(2)} \, x_{\pi(n-1)}\, x_{\pi(n)}$. Hence,
  SGD will quickly converge to predict using the simpler rule, and
  from there on, it would require significantly more gradient updates
  to learn the more complex rule. For more details, see the
  derivation in \cite{shalevsafety}.
\end{remark}

Before we continue, we briefly describe how to boost this negative
result, leading SFT to err almost all the time, while keeping our
diligent learner at a high accuracy. Let us simply concatenate $n$
different copies of the above example, each one relies on its own
permutation $\pi_j$. That is, the input is $x \in \{\pm 1\}^{2n,n}$,
where each $bit$ is generated independently at random. The next
element of the chain is, $v_2 = (v_{2,1},\ldots,v_{2,n})$, where
$v_{2,j} = \prod_{i=1}^n x_{\pi_j(i),j}$. Then, for $i > 2$, the
$i$'th step is $v_i = (v_{i,1},\ldots,v_{i,n})$, where
$v_{i,j} = v_{i-1,j}\,x_{\pi_j(i),j}$. The output is the $n$ bits
string of $v_{n-2}$. By extending the same argument as before, SFT
will learn a rule that has a roughly zero error on the output vector
during training time, while it will have a correctness probability of
roughly $2^{-n}$ at inference time. This is an extreme example of a
distribution drift. In the next section we prove that our diligent
learner succeeds to learn this extended example as well.

\paragraph{Lack of search:} 
The second reason that might cause SFT to fail is the \emph{lack of
  search}. An intuitive example for this failure is our running
example in \figref{fig:coprime}. Indeed, at every split of the
depicted tree, it is difficult to know for sure what should be the
next node, but it is easy to predict a small set of options such that
at least one of them is a correct step. This inherent ambiguity would
similarly make SFT predictions uncertain, and then at inference time,
even one wrong step (which is likely to occur), takes the learner to
an unrecoverable wrong reasoning chain.

To formalize this failure, we construct a second family of problems for which
SFT provably fails. In the next section we will prove that our
diligent learner succeeds for this family as well. 

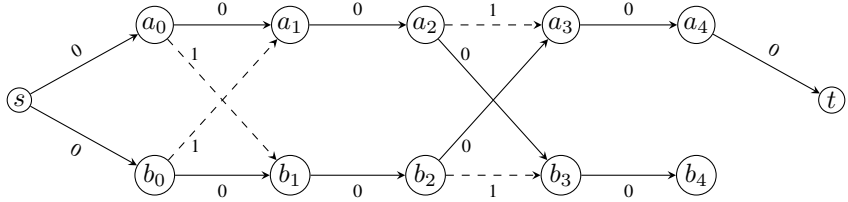
\begin{figure}
\begin{center}
\begin{tikzpicture}[
    >=stealth,
    vertex/.style   = {circle,draw,inner sep=1pt},
    weight/.style={midway,sloped,font=\scriptsize},
    crossWeight/.style={font=\scriptsize},
    weightW/.style={midway,sloped,font=\scriptsize,above}
  ]

  \def\n{4}          
  \def\gap{1.8}      
  \def\W{W}          

  \node[vertex] (s) at (-\gap,1) {$s$};
  \node[vertex] (t) at (\n*\gap+\gap,1) {$t$};

  \foreach \i in {0,...,\n}{
     \node[vertex] (A\i) at (\i*\gap,2) {$a_{\i}$};
     \node[vertex] (B\i) at (\i*\gap,0) {$b_{\i}$};
  }

  \draw[->] (s) -- (A0) node[weight,above]{0};
  \draw[->] (s) -- (B0) node[weight,below]{0};
  \draw[->] (A\n) -- (t) node[weight,above]{0};

  \foreach \j in {2,4}{
     \pgfmathtruncatemacro\jm{\j-1}
     \draw[->] (A\jm) -- (A\j) node[weight,above]{0};
     \draw[->] (B\jm) -- (B\j) node[weight,below]{0};
  }

  \foreach \j/\styleA/\styleB in {1/solid/dashed,3/dashed/solid}{
     \pgfmathtruncatemacro\jm{\j-1}
     \draw[\styleA,->] (A\jm) -- (A\j)
        node[weight,above]{\ifthenelse{\equal{\styleA}{solid}}{0}{1}};
     \draw[\styleA,->] (B\jm) -- (B\j)
        node[weight,below]{\ifthenelse{\equal{\styleA}{solid}}{0}{1}};
     \draw[\styleB,->] (A\jm) -- (B\j)
        node[crossWeight,above,pos=.25]{\ifthenelse{\equal{\styleB}{solid}}{0}{1}};
     \draw[\styleB,->] (B\jm) -- (A\j)
        node[crossWeight,below,pos=.25]{\ifthenelse{\equal{\styleB}{solid}}{0}{1}};
  }

\end{tikzpicture}
\end{center}
\caption{Illustration of the graph.} \label{fig:graph}
\end{figure}

\begin{example} \label{example:search}
Fix an even integer $n\ge 2$.  For every permutation $\pi \in S_n$ and
every input $x=(x_1,\dots,x_n)\in\{\pm1\}^n$ we construct a directed,
edge--weighted graph $G_\pi(x)$ as follows. The vertices are organized
as two horizontal rows of length $n+1$, denoted 
$a_0,\dots,a_n$ and $b_0,\dots,b_n$,
together with a source $s$ and a sink $t$.
Some of the edges are ``static edges'' that always exist. The static
edges are ``start'' edges $(s,a_0)$ and $(s,b_0)$, both of weight $0$, and
an ``exit'' edge $(a_n,t)$ with weight $0$. In addition, there are
input-dependent edges: For every level $j=1,\dots,n$:
\begin{itemize}
  \item If $j$ is \emph{even}, insert the straight edges $(a_{j-1},a_j)$ and $(b_{j-1},b_j)$, both of weight $0$.
  \item If $j$ is \emph{odd}, add the ``straight'' edges
    $(a_{j-1},a_j), (b_{j-1},b_j)$, and add the ``cross'' edges
    $(a_{j-1},b_j), (b_{j-1},a_j)$, where the weights of the edges are
    based on the bit $x_{\pi(j)}$: if it is $1$, then the ``straight''
    edges has a zero cost while the ``cross'' edges has a unit cost,
    and if the bit is $0$ then  the ``straight''
    edges has a unit cost while the ``cross'' edges has a zero cost.
\end{itemize}
An illustration of the graph is given in Figure~\ref{fig:graph}.  Our
goal is to learn to find the shortest path from $s$ to $t$. We model
this as a reasoning chain as follows: $x = v_1 \in \{\pm 1\}^n$ is the
input (and this input uniquely determines the graph). The rest
of the chain describes a shortest path from $s$ to $t$ as follows. For
$i \in \{2,\ldots,n+2\}$, $v_i$ is $+1$ if the
shortest path passes through $a_{i-2}$ and $v_i$ is
$-1$ if the shortest path passes through $b_{i-2}$. Finally,
$y = v_{n+3} = (v_2,v_3,\ldots,v_{n+2})$ is a full description of the
shortest path from $s$ to $t$.
\end{example}

In \lemref{lem:parity} (in Appendix~\ref{sec:shortest}) we prove that
the shortest path is unique, and it goes through $a_0$ or $b_0$
depending on the parity of the odd positions in $x$ according to the
order $\pi$. Therefore, learning to predict $v_2$ from $v_1$ in our
chain requires $\Omega(\exp(n/2))$ number of SGD iterations. As to the
rest of the steps, it follows from \lemref{lem:parity} that $v_i$ can
be determined based on $v_{i-1}$ by the simple rule of ``always pick
the outgoing edge with a zero weight''. This rule can be learnt
quickly by SGD, and therefore SFT would converge to it. And all in
all, the error rate of SFT would be roughly $1/2$.

As with the previous examples, we can boost the failure by
concatenating $n$ different random instances of this graph, while
connecting the target node of the $i$'th instance to the source node
of the $(i+1)$'th instance with a zero weight. This reduces the
accuracy rate of SFT to $2^{-n}$.

This problem seems similar to the previous problem (in
Example~\ref{example:drift}), in the sense that $v_2$ is a parity of
order $n$ bits of the input and the rest of the bits follow from few
previous bits by a simple rule. However, there is a crucial
difference---in the previous problem, the output $y$ \emph{could} be
learnt efficiently from the input by SFT, and the root cause of the
failure of SFT is the distribution drift. In contrast, for the
shortest path problem, the output is the entire shortest path, and
there is no simple way to calculate it from the input. In fact, the
primary cause of failure here is the absence of
\textbf{search}. Basically, it is difficult to know if the first node
in the shortest path is $a_0$ or $b_0$. But given the first node, it
is easy to learn to follow the optimal policy, until we either reach
$a_n$ or $b_n$. Now, in hindsight, if we've reached $a_n$ we know that
our initial decision was good, while if we've reached $b_n$, we know
that our initial decision was bad, and we need to \textbf{backtrack}
to $s$ and try the other option. This example underscores the power of
search. Note that even if we concatenate many copies of the graph,
learning to backtrack to the beginning of each component is an easy
task. The lesson from it is that without embedding backtrack
capabilities explicitly in our learning process, problems might be too
difficult to handle.~\footnote{See also the examples and discussion in
  \cite{bachmann2024pitfalls}.}

\paragraph{Search Tree:} It is convenient to think of the reasoning
process as building a \textbf{search tree}, similarly to the one
depicted in \figref{fig:coprime} for our running example. The root of
the tree is the problem description. Each root-to-leaf path in the
tree is an attempt to build a CoT. If the path ends with a correct
solution to the problem, then we refer to this path as a
\textbf{golden-path} in the search tree. For example, in
\figref{fig:coprime}, the path $0-6-9-10-11-12$ is a golden
path. Clearly, a golden path is a correct and complete CoT. In
contrast, a root-to-leaf path that doesn't end with a correct solution
to the problem is an incorrect CoT, and the optimal backtracking from
it is to the latest ancestor which belongs to some golden path.

It is important to emphasize that the data we are learning from
doesn't contain failed attempts. Each training example only contains a
correct and complete CoT. In the language of the search tree, each
example is a golden-path in some search tree. 

It is tempting to hope that we can create training data of search
trees rather than just having golden-paths. However, this requirement
is unrealistic in many scenarios, and for sure would significantly
limit the amount of data we can learn from. Think for example of
mathematical or algorithmic text books. They contain step-by-step
reasoning chains that prove theorems and lemmas, solve exercises, and
derive algorithmic solutions to problems. But, they rarely contain
failed attempts. And, even if there are some failed attempts, they
would be far from being a complete picture of all possible failed
attempts a learner might encounter. Applying SFT on such incomplete
search trees might still lead to distribution drift problems. For
these reasons, we stick to the assumption that we only have CoT, or
golden-paths, available for training.

\paragraph{Tree-of-Thoughts (ToT):}
\cite{yao2023tree} proposed the Tree of Thoughts (ToT) framework,
attempting to model problem solving as building a search tree over the
combinatorial space of ``thoughts'', where each thought is a coherent
piece of text representing an intermediate step of the solution. As
noted by \cite{yao2023tree}, characterizing problem solving as search
through a combinatorial space goes back to the early work of
\cite{newell1959report}.

The ToT framework relies on prompting LLMs to be an \emph{actor}, that
\emph{generates} nodes of the search tree, as well as a \emph{critic},
that \emph{evaluates} the progress each node makes towards solving the
problem. The third component in the ToT framework is
\emph{search}. To explore the search tree, two variants are proposed: (i)
\textbf{breadth-first search (BFS)} and (ii) \textbf{depth-first
  search (DFS)}.

The BFS approach explores a wide array of shallow paths before
committing to deeper explorations. While such a strategy is
well-suited for certain structured games or planning tasks, it poses a
significant limitation in our setting. In many reasoning
problems---especially those involving uncertainty, ambiguity, or
delayed resolution of correctness--it may be impossible to evaluate
the quality of a reasoning path without venturing deep enough to
gather conclusive evidence. In Appendix~\ref{sec:failure_BFS_TOT} we
prove that ToT-BFS would fail on the boosted version of
Example~\ref{example:drift}, where an incorrect choice made early on
may only be revealed as such after several subsequent steps. For this
reason, the critic might not be able to learn to separate different
options soon enough.

Consequently, ToT-BFS may be forced to maintain and explore an
exponentially large number of shallow reasoning paths before it
accumulates enough hindsight to rule out incorrect trajectories. This
leads to a combinatorial explosion in the number of paths under
consideration, rendering the algorithm computationally infeasible for
problems that require significant depth to verify correctness. This is
also evident by the experiments in \cite{yao2023tree}, where ToT-BFS
was used for problems that require a very shallow search tree (depth
of 3 for the ``Game of 24'' problem and depth of 2 for ``creative
writing''). But for the the third problem considered in
\cite{yao2023tree}, the ``mini crosswords'' problem, they switched to
using ToT-DFS since the problem requires a depth $10$ reasoning
chains.

Depth-first strategies that incorporate backtracking can more
efficiently isolate incorrect reasoning paths by allowing the model to
follow a single trajectory until failure is detected, and then revise
its course. However, DFS might also suffer from an exponentially large
number of visited nodes. Indeed, in Appendix~\ref{sec:failure_DFS_TOT} we
prove that the ToT-DFS of \cite{yao2023tree} requires an exponentially
large number of visited nodes for the boosted version of
Example~\ref{example:drift}. The missing ingredient is a learnable
backtracking strategy---ToT-DFS always backtracks to the parent
node. This misses the observation of delayed resolution of
correctness, where once we explore enough, we may be able to
understand that the whole direction is wrong and backtrack to an
earlier node than the parent node. As an example, observe the
backtrack in \figref{fig:coprime} from node 5 all the way to node 0,
since at that point, the model can understand that an entire proof
attack is wrong.

These critical observations motivate the design of our diligent
learner, which prefers DFS over BFS and explicitly learns mechanisms
for backtracking and correction, allowing it to navigate reasoning
trees more efficiently than approaches such as ToT.

\paragraph{Monte-Carlo-Tree-Search (MCTS):}

MCTS is a widely used algorithmic framework for reasoning and
decision-making under uncertainty.  It operates by simulating many
possible continuations of a decision path in order to guide future
choices. Each simulation has 4 steps: selection, expansion,
evaluation, and backpropagation. Variants of MCTS differ in how
they perform these 4 steps. We focus on the variant described in
\cite{kim2025astro}, as it is tailored to reasoning problems.

The \emph{selection} is based on the PUCT (Predictor + Upper
Confidence Tree) score, which was proposed by
\cite{silver2017mastering} for the AlphaGo algorithm. The score
combines a learnt Q function, the probability assigned by the LLM to
the next node, and the visit count of the state. For the
\emph{expansion} step, they sample several possible reasoning steps
from the LLM, and for each one, they rollout (based on the selection
score) until reaching a full solution. Then, they \emph{evaluate} the
node based on the outcome reward of the rollouts.  Finally, in the
\emph{backpropagation } phase they update the visit counts of the
nodes and also update the $Q$ function based on the visit counts and
the evaluated reward of the nodes.

The main weakness of the MCTS approach is that it also might require
an exponentially large number of steps to converge. To see this, think
again of the boosted versions of Example~\ref{example:drift} or
Example~\ref{example:search}. In both versions, the probability to
reach a valid outcome is exponentially small (in $n$). It follows that
the rollouts of the MCTS have a very low probability of obtaining a
reward signal for learning.  Consequently, MCTS might also be forced to
maintain and explore an exponentially large number of shallow
reasoning paths before it accumulates enough hindsight to rule out
incorrect trajectories.

\paragraph{Reinforcement Learning (RL) with unstructured exploration:}
We next consider RL methods that allow the model to explore
freely. Two popular algorithms are PPO~(\cite{schulman2017proximal})
and GRPO~(\cite{shao2024deepseekmath}), where in the context of Large
Reasoning Models (LRMs), the hope is that in-context search would
emerge automatically just from trial and error.  This in-context
search may learn to traverse over a search tree in words. For example,
for the tree in \figref{fig:coprime}, after repeating the question
(node 0 in the tree), the model can elicit the text in nodes
1, 2, 3, 4. Then, node 5 in the tree (the backtrack node) can be
described as ``we are stuck, so let's try another direction''. Then
comes the text of nodes 6, 7. Then, node 8 (another backtrack node)
can be described as ``the value of $k=2$ doesn't work, so let's try
another value''. And finally, the text of nodes 9, 10, 11, and 12
leads to the desired result.

For concreteness, let's describe how the GRPO algorithm would work in
our context. The training starts with vanilla SFT on the CoT examples.
Then, each RL iteration involves sampling $B$ generations of a full
CoT that ends with a proposed solution to the problem. In these
generations, the model may be encouraged 
to perform search operations like criticizing itself (using the
buzzword ``wait'') followed by backtracking (``let's go back to step
X'') or reset (e.g. using the buzzword ``I should start over''). Then,
an outcome reward determines which of the $B$ generations ended up
with a correct solution and which failed. An ``advantage'' estimation is
calculated per each generation (where in our case, it is the reward
for the generation minus the averaged reward over all generations,
possibly with some normalization). And finally the model weights are
updated based on these advantage estimations.

The problem with these approaches is that if the initial probability
(after the cold-start SFT) to generate a succesful CoT (one that ends
with a correct answer) is much smaller than $1/B$, then the advantage
function will become zero, because all of the generations are bad, and
we don't have a signal for learning. Similar problem holds for any
method that performs unstructured exploration and has a very small
probability to succeed. This is particularly problematic in reasoning
problems, where we may have many reasoning steps and a fixed
probability to fail in each one of these causes the probability to
generate a full succesful CoT to be exponentially small.

It follows that if we apply such an RL algorithm on the boosted
version of Example~\ref{example:search}, then we will not have any
signal for learning (unless exploring exponentially large number of
times).  A possible remedy is to ``shape'' the reward, e.g. by giving
a positive reward to backtrack from nodes which are not on the golden
path to the golden path. However, this is tricky since nodes that are
not part of the golden path may be equivalent to nodes which are on
the golden path (say, the same logical reasoning step with a different
phrasing). In this case, the shaped reward might give wrong
reinforcement signals. Furthermore, a flat reward from every off-path
to golden-path backtracking might push the model to ``give-up'' too
early, and there may be no ``simple rule'' that manages to
understand we are off the path at early stages.

\paragraph{A teacher generates search trees for a student:}
\cite{kim2025astro} recently introduced a framework for training
language models to reason like search algorithms called ASTRO. The
idea is to generate search trees using MCTS, and then apply SFT+RL, as
described in previous paragraphs. They demonstrate performance gains
on some standard benchmarks, and in particular show that
search-inspired training instills reasoning capabilities. This message
is well aligned with our approach. However, as we have shown
previously, the MCTS variant that they propose is not well suited for
difficult reasoning problems. In particular, for problems that MCTS
fails to generate good solutions (such as the ones we've shown
previously), ASTRO would fail as well. In addition, ASTRO might suffer
from a distribution drift problem---just like the Anna Karenina
Principle, ``All happy families are alike; each unhappy family is
unhappy in its own way''---the mistakes of the teacher are not
necessarily aligned with the mistakes of the student, and therefore,
the training data that contains recovery from the teacher's student is
not necessarily sufficient for learning to recover from the student's
unique mistakes.

\paragraph{Reverse Curriculum Reinforcement Learning:}
\cite{xi2024training} proposed the ``Learning Reasoning through
Reverse Curriculum RL'' ($R^3$) algorithm. The motivation of $R^3$ is
the shortcoming of the GRPO algorithm we have outline previously---if
the probability to generate a correct reasoning chain is very small,
then outcome reward alone does not give us any signal for learning.
To overcome this problem, $R^3$ applies a reverse curriculum over the CoT
examples. Initially, it shows all the CoT except the last step as a
prefix, and use a policy-gradient RL method (PPO) to learn to complete the
sequence while only relying on outcome reward. Then, it gradually
increases the number of steps that the RL should learn, until the last
phase of learning in which the RL only observes the model and should
generate the entire CoT sequence. \cite{xi2024training} explains the
rational behind the reverse curriculum as follows: ``By slowly moving the
start state from the end of the CoT to the beginning, the model faces
an easy exploration problem at each point where it is likely to
succeed, since it has already learned to solve most of the remaining
parts. In this way, a curriculum of gradually increasing exploration
difficulty is created, and we can provide approximately step-by-step
supervisory signals for the model.'' A similar approach for playing
difficult games has been proposed in \cite{salimans2018learning}, and 
recently, \cite{amani2025rl} proposed a variant of this approach, in
which the part of the CoT that is being observed during training is
adapted automatically.

Our diligent learner also applies a reverse curriculum in order to
control the success of exploration. But unlike $R^3$ and its
adaptation in \cite{amani2025rl}, we also explicitly add \emph{search}
components to the model, in particular an explicit training of how to
backtrack. Our training approach does not rely on policy-gradient but
instead relies on explicit exploration in order to generate examples
for candidate step creation and backtracking. To see why search is
crucial, consider again Example~\ref{example:search}, and suppose that
we apply the $R^3$ algorithm on it. Due to the reverse curriculum, in
the first steps of the algorithm it is likely to succeed in learning
the simple rule of ``always pick the outgoing edge with a zero
weight''. The reverse curriculum would succeed and then we will get to
the point where we only show $v_1$ (the graph itself) and the
algorithm needs to decide whether the first step of a shortest path
should pick node $a_0$ or $b_0$. The initial probability to pick any
of this option is roughly $1/2$. And, due to the curriculum learning,
once the initial node is picked, for the following nodes the network
would continue to pick successor nodes for which the outgoing edge is
with a zero weight. In roughly half of the time this would lead us to
$a_n$ and we get a positive outcome reward. In the other cases, we
would reach $b_n$. From there, nothing tells us what to do (the
network never saw this case previously). It may very well be the case
that the network would stop there, or output $a_n$ even if its
illegal. In those cases, the network would get a negative
result. Then, the gradient propagation tells the model to prefer the
good examples over the bad examples. But this practically gives the
model additional examples for figuring out the parity problem. Since
learning parity takes exponential iterations, we would not learn
anything new from these examples. The only way that the algorithm
would learn something useful for this task if it happens to ``guess''
that a backtrack is required, and it also happens to guess where does
it have to backtrack to, and how to elicit a final correct shortest
path answer after erring and backtracking. There is no good reason to
assume that this behavior would spontaneously emerges, even in this
rather simple example.

To summarize, existing algorithms lack in at least one of the
following components: (1) distribution drift (2) lack of search (3)
exponential blowup at train and/or inference time.
In the next section we describe and analyze our \emph{diligent
  learner}, that explicitly tackles these three components.

\section{The Diligent Learner: An Efficient Learner from CoT Data} \label{sec:diligent}

The \emph{diligent learner} is an algorithm whose output is a
specialized auto-regressive LLM that explicitly builds a search tree
for problem solving, similarly to the one given in
\figref{fig:coprime}.

We first need to formally describe what is a search tree and how an
auto-regressive LLM generates one. A search tree for reasoning is a
rooted tree, where each node is associated with a text chunk in
$\Sigma^*$ that represents a coherent paragraph of reasoning.  The
root node is the problem description. Nodes of the tree are assigned
identifying labels. For simplicity, let's say that labels are numbers
in $\{0, \ldots, N-1\}$, where $N$ is the number of nodes in the tree,
the root node is assigned the label $0$, and the rest of the nodes are
assigned numbers according to their creation order. But any other
labeling system can work as well.  Each root-to-leaf path in the tree
is a proposed CoT. Leaves are either designated as a ``done'' leaf or
as a ``backtrack'' leaf. If the leaf is designated as a ``done'' leaf,
it means that the path to this leaf is a proposed complete CoT for the
problem, and the leaf node describes the proposed solution to the
problem. When a ``done'' leaf is generated, the tree construction
terminates. If the leaf is designated as a ``backtrack'' leaf, it
should specify the label of the node to which we backtrack from the
leaf. This target node of the backtracking operation must be a node in
the path from the root of the tree to the current backtrack leaf.

An auto-regressive LLM defines a distribution,
$\pi_\theta(t_i|t_{i-1},\ldots,t_0)$, of generating the token $t_i$
given previous tokens $t_{i-1},\ldots,t_0$. The weight vector $\theta$
is the vector of parameters of the model, which we should learn.  We
use \emph{constrained decoding} in order to force our LLM to generate
text that describes
node-to-leaf paths in a search trees.~\footnote{See for example the
  library llguidance in
  \url{https://github.com/guidance-ai/llguidance}} This is done by
relying on a few special tokens:
\begin{itemize}
\item \texttt{<node>} label text \texttt{</node>} \\
  This describes a node of the tree, with an identifying label and with
  text that describes the reasoning step.
\item \texttt{<backtrack>} label \texttt{</backtrack>} \\
  This means that we add a backtrack leaf node, traversing back to
  the node identified by the label, so that the next node-to-leaf path we would
  generate will start from that node. We can only backtrack to a
  node on the path from the root of the tree to the leaf.
\item \texttt{<done>} text \texttt{</done>} \\
  This creates a ``done'' leaf node, with the solution to the problem.
\end{itemize}

The grammar of a node-to-leaf path is simple: add a sequence of nodes,
each one is a child of the previous node, until generating either a
``backtrack leaf'' or a ``done leaf''. Forcing the LLM to adhere to
these rules is done by setting the probabilities of invalid tokens to
zero and re-normalizing.

Once we have a mechanism to generate a node-to-leaf path, generating a
full search tree is done by sequentially generating node-to-leaf
paths as follows. If the previous path ended with a backtrack leaf,
then the next generation starts from the node to which the backtrack
leaf points to. And if the previous path ended with a done leaf, we
terminate the construction of the search tree.

\begin{remark}
  When we generate a node-to-leaf path from a given node, there are
  two possible contexts that one may consider:
  \begin{enumerate}
  \item The context is the path from the root of the tree to the start
    node of the path. This is the configuration we analyze in this work. From the
    implementation perspective, this leads to efficient processing
    by key–value (KV) caching.
  \item The context is the entire search tree that was generated up
    until now. This is the approach that is taken by some in-context
    LRM models that were trained by RLVR as well as the ASTRO
    algorithm mentioned in the previous section. While this approach
    has the benefit of ``telling'' the model about previous
    unsuccessful attempts, we do not recommend it as it might cause a
    distribution drift (this will become evident once we will explain
    how the learning process is performed).
  \end{enumerate}
\end{remark}

\begin{remark}
  Since the model is constrained to generate tokens that has a
  specific structure, we can use a validation model that checks the
  logical soundness of each transition in the path from the root node
  to the ``done'' node. As discussed previously, validation is much
  easier than generation. The validator only needs to check that each
  node follows logically and factually from the previous nodes. See
  the discussion in \cite{shalev2024artificial}.
\end{remark}

Our goal now is to learn a parameter vector $\theta$ that leads to
``good'' generation of search trees, in the sense that the tree
creation is terminated fast enough and the result is correct according
to the reward function.  We start with stating sufficient properties
of $\theta$ that leads to a successful creation of a search tree.

  We consider problems for which the maximal length of a correct
  reasoning chain is strictly smaller than $T_{\max}$ and so we force
  a backtrack node whenever we reach depth $T_{\max}$ in the tree.
  The procedure we have described builds a search tree in a
  depth-first manner, where the LLM, $\pi_\theta$, chooses at each
  node whether to generate a new node or a leaf.  The following lemma
  provides sufficient conditions for a successful search tree
  creation. These conditions are closely related to our sufficient
  conditions for CoT learnability.
\begin{lemma} \label{lem:success}
  Let $\gamma \in (0,1)$ and $\delta \in (0,1/2)$ be
  constants. Define
  \[
    \epsilon = \frac{\gamma\,\delta}{2T_{\max}} ~~~\textrm{and}~~~
    B = \left\lceil \frac{\log(T_{\max}/\delta)}{\gamma} \right\rceil ~.
  \]
  Suppose that we run the search tree creation while restricting at
  most $B$ attempts to generate a child of each node.
  Assume that the following holds for every $i < T_{\max}$:
  \begin{itemize}
  \item With probability of at least
    $1-\epsilon$ over sequences $v_1,\ldots,v_i$ which are correct but
    incomplete (where the probability is the one induced by
    $\pi_\theta$), we have that generating a correct continuation node
    $v_{i+1}$ based on $\pi_\theta$ is $\gamma$-correct (as defined in
    \eqref{eqn:gammaCorrect}).
  \item With probability of at least
    $1-\epsilon$ over sequences $c = (v_1,\ldots,v_{T_{\max}})$, which are
    incorrect and for which $\beta(c) = i$, we have that $\pi_\theta$
    will induce a backtrack leaf node to $i$ (at some node $j \in
    \{i+1,\ldots,t_{\max}\}$)
  \end{itemize}
  Then, the probability that $\pi_\theta$ would generate a correct CoT
  is at least $1-4\delta$.
  Furthermore, the number of backtrack leaves in the tree is at most
  $(B-1)\,(T_{\max}-1)$. 
\end{lemma}
\begin{proof}
We begin the construction from $v_1$. The probability to succeed to reach $v_2$ is
calculated as follows. We have a probability of at least
$(1-\epsilon)$ that $\pi_\theta$ is $\gamma$-correct for generating
$v_2$. Then, given it is $\gamma$-correct, we have $B$ attempts. If we
succeed at attempt $i$, it means we failed at $i-1$ attempts, but
we've succeeded to backtrack correctly at each failed attempt. Summing
over $i$ we obtain the following lower bound on the probability to
succeed to reach $v_2$:
\begin{align*}
&    (1-\epsilon)\,\sum_{i=1}^{B} \left[(1-\gamma)
  (1-\epsilon)\right]^{i-1} \gamma 
    ~=~ (1-\epsilon)\, \gamma \sum_{i=0}^{B-1} \left[(1-\gamma)
  (1-\epsilon)\right]^{i} \\
    &= (1-\epsilon) \gamma \frac{1 -
      (1-\gamma)^B(1-\epsilon)^B}{1-(1-\gamma)(1-\epsilon)} \\
  &\ge (1-\tfrac{\epsilon}{\gamma})\,(1 -
    (1-\gamma)^B(1-\epsilon)^B)
\end{align*}
where in the last inequality we used \lemref{lem:geinq} from
\secref{sec:technical}.  Next, using the inequality $1-x \le e^{-x}$
that holds for every $x$, we have that the above is bounded by
\[
\ge (1-\tfrac{\epsilon}{\gamma})\,(1 - e^{-(\gamma+\epsilon)\,B})
\]
Using the definition of $B$, the above is bounded by
\[
\ge (1-\tfrac{\epsilon}{\gamma})\,(1 - \tfrac{\delta}{T_{\max}})
\]
Next, we use the inequality $e^{-2x} \le 1-x$ which holds for $x \in
[0,1/2]$ to obtain that the above is lower bounded by
\[
\ge e^{-2\tfrac{\epsilon}{\gamma} - 2\tfrac{\delta}{T_{\max}}}
\]
So, with the above probability we will reach a correct
$v_2$. Repeating this process $T$ times, where $T < T_{\max}$ is the required
length of a correct reasoning chain, we obtain that the probability to
succeed is lower bounded by
\[
\ge e^{-2\tfrac{\epsilon}{\gamma}\,T - 2\tfrac{\delta}{T_{\max}}\,T}
\]
Using the definition of $\epsilon$ this is lower bounded by
\[
\ge e^{-4\delta} \ge 1 - 4\delta ~,
\]
where as used earlier, we apply the inequality $e^{-x} \ge 1-x$. 
\end{proof}

\begin{remark}
  \lemref{lem:success} assumes a $(1-\epsilon)$ probability to
  backtrack to the exact right node, where $\epsilon$ is rather
  small. We can significantly relax this assumption as follows. Let
  $c = (v_1,\ldots,v_T)$ be some incorrect sequence. Then, we can
  allow the backtrack to have a fixed probability to undershoot,
  namely, to return $i$ s.t. $i \in \{\beta(c)+1,\ldots,T-1\}$. It can
  be shown that in this case, we can still guarantee fast convergence
  with an averaged constant factor of additional backtrack
  leaves. However, overshoot errors (in which $i < \beta(c)$) should
  have a small enough probability. Intuitively, as in the popular
  Chutes and Ladders game, after working hard and progressing in the
  search tree, an overshoot of the backtrack may revert the search to
  a significantly earlier state. Since undershoots can be tolerated,
  we can tune the backtrack operation so it would have a very small
  probability of an overshoot at the expense of a higher probability
  to undershoot.
\end{remark}

\subsection{The Learning Process}

The LLM needs to learn to generate ``new'' nodes, ``done'' nodes, and
``backtrack'' operations. At each step of our algorithm, we build a
new training set of examples, in which each example consists of a
context-response pair. The context of each example contains the text
describing the path from the root of the tree to the current node. The
response is a new node, a done node, or a backtrack operation.  The
main challenge is to ensure that the sizes of the trees we build
during the learning process and during inference time do not grow
exponentially in depth.

The algorithm has a set of golden-path examples, denoted $G$. Each
golden-path example in $G$ is a sequence $g = (v_1,\ldots,v_{|g|})$,
where $|g|$ denotes the length of the reasoning chain, $v_1$ is the problem
description, $v_{|g|}$ is a correct solution, and each $v_i$ follows
from $v_0,\ldots,v_{i-1}$. Our diligent learner employs an inductive
learning. Conceptually, we are going to learn a series of functions,
$f_1,f_2, f_3,\ldots$. Each $f_t$ assumes that it gets a prefix of a
reasoning chain that can be completed to a correct reasoning chain in
exactly $t$ steps. Internally, $f_t$ builds a search tree that starts
with a single reasoning chain and at some point starts to branch out,
but from the first branching point the depth of the subtree is exactly
$t$ and the number of leaves is $O(t)$. Crucially, the size of the
tree isn't exponentially large in $t$. When learning the function
$f_{t+1}$ we would rely on $f_t$, and by doing so we prevent an
exponential blowup.

This reverse curriculum approach is similar to the $R^3$ algorithm of
\cite{xi2024training}. However, unlike $R^3$, which does not
explicitly embed search and relies on policy-gradient, we are using a
reduction to supervised learning. That is, at each step of our
algorithm we augment the training set of examples with more examples
that are obtained by exploration. This is similar to the approach
taken by the DAgger RL algorithm of~\cite{ross2011reduction}, but the
major difference is that unlike DAgger, we do not assume the existence
of an expert oracle for the optimal action which can be queried at
every visited state.

The algorithm starts with the creation of $f_1$. This is done as
follows. For every golden path $(v_1,\ldots,v_{|g|}) \in G$, we let
the context be $c = (v_1,\ldots,v_{|g|-1})$ and we add the example
($c, $ \texttt{<done>} $v_{|g|}$ \texttt{</done>}) to our training set
as an example for a ``done'' node creation. We then perform SFT on
these examples. Next, we explore by generating $B$ candidates,
$v^{(1)}_{|g|}, \ldots, v^{(B)}_{|g|}$, from the resulting
$\pi_\theta$. For each such candidate, if it satisfies the edge
function, we calculate the outcome reward function. If the reward is
$1$, we add an example ($c, $ \texttt{<done>} $v^{(i)}_{|g|}$
\texttt{</done>}), as an example for a ``done'' node creation. If the
reward is $0$, we add an example ($c, $ \texttt{<backtrack>}
id($v_{|g|-1}$ )\texttt{</backtrack>}), where id($v_{|g|-1}$) is the
label of the node $v_{|g|-1}$ in the tree. This is an example of a
``backtrack'' operation. Finally, we perform SFT on the updated
training set. The examples we have constructed demonstrate both a
``done'' node creation and backtracking of one step backward. If the
learning succeeds (in the sense of satisfying the conditions in
\lemref{lem:success}), then we have managed to successfully create the
function $f_1$, and according to \lemref{lem:success}, the number of
leaves we would create is $O(1)$. Note that under our assumptions on
CoT learnability, the learning would succeed to guarantee the
conditions of \lemref{lem:success}, and therefore the first step of
our construction works.

Next, consider the inductive step. Suppose that we have generated the
functions $f_1,\ldots,f_t$ and now we aim at learning $f_{t+1}$. We
use every golden path $(v_1,\ldots,v_{|g|}) \in G$ such that
$|g| > t+1$. The context is $c = (v_1,\ldots,v_{|g|-t-1})$ and we
first add the example ($c, $ \texttt{<node>} $v_{|g|-t}$
\texttt{</node>}) to our training set as an example for a new node
creation. We perform SFT with these examples. Next, we explore by
generating $B$ candidates, $v^{(1)}_{|g|-t}, \ldots, v^{(B)}_{|g|-t}$,
from the resulting $\pi_\theta$. By our inductive assumption, for
every $i \in [B]$, if $v^{(i)}_{|g|-t}$ can be extended to a correct
reasoning chain of length $|g|$, then generating a sub-tree by calling
$f_t(v^{(i)}_{|g|-t})$ would generate $O(t)$ leaves before reaching a
correct solution. Therefore, for every $i$, we stop the generation of
the sub-tree after it creates $O(t)$ leaves. This is crucial as
otherwise, an incorrect $v^{(i)}_{|g|-t}$ might create a tree of
exponential size (as our inductive guarantee for $f_t$ only holds if
we start with a correct prefix).  All in all, we would generate
$O(tB)$ leaves. Now, for every $i$, if the sub-tree generated by $f_t$
reached a correct solution, then we add all of the nodes on the path
to the correct solutions as examples for the new node (or done node)
creation, and we add every unsuccessful leaf as a backtrack example to
the path to the correct solution. In addition, for every leaf in every
$i$ that did not yield correct solutions, we add a backtrack example
to $v_{|g|-t-1}$. We again perform SFT on the updated training set of
examples. By our assumptions on CoT learnability, the learning process
would succeed. It follows that for a wrong leaf, there is at least a
$(1-\epsilon)$ probability to backtracking to the correct node. We can
now apply \lemref{lem:success} to get that $f_{t+1}$ will create a
tree with $O(t+1)$ leaves before succeeding, and this concludes the
inductive construction.

\begin{remark}
  It is not hard to see that the conditions of \thmref{thm:main} hold
  for Example~\ref{example:drift} and
  Example~\ref{example:search}. Hence, our diligent learner succeeds
  in learning these problems.
\end{remark}

\section{Discussion}

This work investigates the fundamental limitations of existing
learning algorithms in leveraging Chain-of-Thought (CoT) data and
introduces a new learning algorithm---The Diligent Learner---designed
to overcome those challenges. Through formal analysis and synthetic
examples, the paper identifies three critical bottlenecks that
undermine current methods: distribution drift, lack of embedded
search, and exponential growth in computational cost at inference or
training time.

The proposed Diligent Learner provides a principled approach to CoT
learning under two realistic assumptions: (1) the availability of a
generative model that produces correct next reasoning steps with some
non-zero probability ($\gamma$-GPAC learnability), and (2) the ability to
backtrack efficiently upon detecting a failure in the reasoning
path. These assumptions are significantly less restrictive than those
in prior theoretical frameworks, which often unrealistically assume a
deterministic token-level prediction.

Importantly, the analysis demonstrates that popular algorithms---SFT,
MCTS, ToT, PPO-based RL, and reverse curriculum approaches like
R3---fail on carefully constructed problems due to their inability to
efficiently handle the inherent uncertainty and ambiguity of reasoning
tasks. The Diligent Learner addresses these issues by explicitly
modeling reasoning as a depth-first search over a tree of reasoning
steps, integrating backtracking, and controlling exploration depth via
reverse curriculum.

From a practical standpoint, this approach aligns closely with how
humans solve complex reasoning tasks: iteratively exploring paths,
recognizing when a line of thought is unproductive, and backtracking
to try alternatives. In contrast, existing machine learning techniques
often behave as brittle mimics, unable to recover from deviation and
easily misled by shortcut signals in the training data.

While the theoretical guarantees provided are strong and illustrative,
it remains to validate their practicality in natural domains where
reasoning is essential. A key challenge lies in scaling the approach
to large models and empirically validating these assumptions on
real-world CoT data..

In conclusion, this paper lays a rigorous foundation for learning to
reason from CoT data in a more human-like, resilient, and structured
manner. It calls for a reevaluation of how search, supervision, and
curriculum interact in reasoning-capable models, and paves the way for
building Large Reasoning Models (LRMs) that go beyond pattern matching
and toward genuine problem-solving abilities.

\appendix

\section{Multiplication is Hard for SGD Learning} \label{sec:multiplication}

Using Auto-regressive transformers for learning multi-digit multiplication
involves feeding as input two strings of digits $a,b$, representing
integers $N(a),N(b)$ in the usual decimal basis, and the network
should output a string of digits forming the decimal representation of
$N(a) \cdot N(b)$, one digit at a time. We build a family of
``simpler-multiplication'' problems. The first step of the vanilla
long multiplication algorithm for calculating $N(a) \cdot N(b)$
involves different summation of $a_ib_j$ over pairs $(i,j)$ followed
by moduls 10. In our ``simpler-multiplication'' operation, the input
is again two strings of digits and the output is a single digit
representing $\sum_{i=0}^{m-1} a_i b_{m-1-i} ~\bmod~ 10$. This is
exactly the $m$-th digit of $N(a) \cdot N(b)$ if we ignore the second
step of long multiplication in which a carry over from lower digits is
also summed (modulo 10). In this sense, ``simpler-multiplication'' is
indeed simpler than the full multiplication solution. Each function in
our family of ``simpler-multiplication'' problems involve a different
fixed permutation of the order of digits in the input. We prove that
SGD requires exponentially many steps to learn functions in this
family.

\subsection{Analyzing SGD convergence via Low Variance}

Analyzing SGD convergence via low variance has been studied in the
past (e.g. as in \cite{malach2022hardness}), but these results were
done for binary classification problems. Our purpose in this section
is to extend these results to next-token prediction.

Let $\Sigma$ be a finite alphabet (our dictionary of tokens) of size
$k$.  Let $h_w(x)$ be a probability of next token from $\Sigma$ given
a context $x$, modeled as a vector
$[\log(p_w(x))[1],\ldots, \log(p_w(x))[k]]$ where $k = |\Sigma|$. For
example, $w$ are the weights of a transfomer network, $x$ is the
context string, and the network produces a probability vector,
$p_w(x) \in [0,1]^k$, where $p_w(x)[i]$ is the probability that the
next token will be the $i$'th token in $\Sigma$ given the context $x$.

Let $f : X \to [0,1]^k$ be a target conditional probability function we aim to learn, that is, $f(x)$ is a probability vector of the true underlying probability of the next token given the context $x$. We are using the input specific loss function
\[
\ell(h_w(x),f(x)) = -\inner{h_w(x),f(x)} = \E_{y \sim f(x)} -\log(p_w(x)[y])
\]
as the standard log loss function
(which equals to the KL divergence between $f(x)$ and $p_w(x)$ up to a constant).
Our loss function is the average over $x \sim D$, 
\[
  L_f(h_w) = \E_{x \sim D} \ell(h_w(x), f(x)) =
  - \E_{x \sim D} \inner{h_w(x),f(x)} = -\inner{h_w,f}_D
\]
where for two functions $g_1,g_2$ both from $X$ to $\reals^k$ we
define
\[
  \inner{g_1,g_2}_D = \E_{x \sim D} \inner{g_1(x),g_2(x)} ~. \]
Since
the range of $h_w$ is $\reals^k$, its gradient with respect to $w$ for
a fixed $x$ is a matrix $\nabla_w h_w(x) \in \reals^{k,d}$ where $d$
is the dimension of $w$. By linearity of the gradient, the gradient of
$L_f(h_w)$ w.r.t. $w$ is
\[
  \nabla_w L_f(h_w) = - \E_{x \sim D} \nabla_w \inner{h_w(x),f(x)} =
  - \E_{x \sim D} (\nabla_w h_w(x))^\top f(x)  \in \reals^d~.
\]
We will apply the common practice of clipping all the elements of the
gradient matrix by a constant, and denote $\nabla_w^c h_w(x)$ to be
the matrix whose $(i,j)$ element is the $(i,j)$ element of
$\nabla_w h_w(x)$ clipped to the regime $[-1,1]$. We use the notation
$\nabla^c L_f(h_w) = - \E_{x \sim D} (\nabla_w^c h_w(x))^\top f(x)$.

In our learning problem, $f$ comes from a family $\mathcal{F}$.
Define $\bar{f}$ to be the constant function that outputs for every
$x$ the uniform distribution over $\Sigma$, that is, for every $x$, $\bar{f}(x) =
[1/k,\ldots,1/k] := U_k$.
We define the variance~\footnote{In the typical definition of a
  variance, $\bar{f}(x)$ refers to 
  $\E_f f(x)$ for every $x$. While we use $\bar{f}$ to be the constant
  function that outputs $U_k$ for every input, we still allow
  ourselves to adopt the terminology of
  ``Variance''.}
of the gradient as
\[
  \E_{f \sim \Unif(\mathcal F)}
    \|\nabla^c L_f(h_w) - \nabla^c L_{\bar{f}}(h_w)\|^2
  \]
  We will use $\E_f$ as a shorthand for
  $\E_{f \sim \Unif(\mathcal F)}$. The variance of the gradient
  measures how much we will move
  differently than the constant direction $\nabla^c L_{\bar{f}}(h_w)$
  (which doesn't depend on the specific function $f$ we aim to learn).
  If this quantity is very small, it means that for most of the
  functions $f$ we'll change the weight vector $w$ in almost the same
  manner, meaning that we can't separate learning with labels from one
  function to learning with labels from the other one.

We have
\begin{align*}
  \E_f \|\nabla^c L_f(h_w) - \nabla^c L_{\bar{f}}(h_w)\|^2
  &= \E_f \sum_{i=1}^d \left( - \E_{x \sim D}
    \inner{\nabla_w^c h_w(x)[:,i], f(x)-\bar{f}(x)} \right)^2 \\
  &=
    \sum_{i=1}^d \E_f  ( \inner{\nabla_w^c h_w[:,i], f-\bar{f}}_D)^2
    =
    \sum_{i=1}^d \E_f  ( \inner{\nabla_w^c h_w[:,i], f-U_k}_D)^2
    ~\le~
    d\,\mathrm{Var}(\mathcal{F})
\end{align*}
where we define,
\[
\mathrm{Var}(\mathcal{F}) = \sup_{\phi : X \to [-1,1]^k} \E_f\left[
\left(\inner{\phi, f-U_k}_D\right)^2 \right]~.
\]

\subsection{Bounding the Variance of a family of Multiplication-like
  functions}

Let the alphabet be
\[
  \Sigma=\{0,\dots,9\}, \qquad k\;=\;|\Sigma|=10 .
\]

The input distribution is as follows. Fix an \emph{even} sequence
of length $n=2m$. Draw
\[
   x=(x_{0},\dots,x_{2m-1})\;\sim\; D
   \quad\text{with i.i.d. coordinates}\quad
   x_i\sim\Unif(\Sigma).
\]

\paragraph{The ``no-carry product digit'' target:} 
Split $x$ into two $m$-digit numbers
\[
   a=\sum_{i=0}^{m-1}x_i\,10^{i},
   \qquad
   b=\sum_{i=0}^{m-1}x_{m+i}\,10^{i},
\]
and define the target digit
\[
   g(x)
   \;=\;
   \Bigl(
      \sum_{i=0}^{m-1} a_{i}\,b_{m-1-i}
   \Bigr)\bmod 10
   \;\in\Sigma .
\]
This equals the decimal digit of $a\cdot b$ at position $m-1$ (the
coefficient of $10^{m-1}$) if all carries are zero, so it retains the
flavour of multiplication yet is arithmetically simpler.

We can write $g(x)$ as
\begin{equation} \label{eqn:g_def}
g(x) = \Bigl(\sum_{i=0}^{m-1} x_i x_{2m-1-i}\Bigr)\bmod 10
\end{equation}

\paragraph{The family of label functions:}
For every permutation $\pi \in S_m$ and $x \in \Sigma^{2m}$,
let $x_\pi$ be the string whose first $m$ digits are permuted
according to $\pi$. Let $f_\pi : X \to [0,1]^k$ be defined as:
\begin{equation} \label{eqn:f_def}
f_\pi(x) = U_k + k^{-1}\,(e_{g(x_\pi)} - e_{g(x_\pi)+5 \bmod 10})
\end{equation}
That is, $f_\pi(x)$ is the uniform probability vector
$U_k = [1/k,\ldots,1/k]$, except that we increase position $g(x_\pi)$
from $1/k$ to $2/k$ and we decrease position
$g(x_\pi)+5 \bmod 10$ to be 0.

Our family of functions is 
\begin{equation} \label{eqn:def}
   \mathcal F
   \;=\;
   \bigl\{f_\pi \;:\; \pi\in S_{m}\bigr\}.
 \end{equation}
 Note that the sole difference between different functions in the
 family is the order they interpret the input, so we expect the
 difficulty of learning each member of this family by a transformer to
 be the same. In fact, by this argument, due to symmetry, if one
 member of this family
 is difficult to be learnt, so is all others.

 Our main result is an upper bound on the variance of this family
 \begin{theorem} \label{thm:main_mult}
   For a permutation $\pi \in S_m$ let $f_\pi$ be as defined in
   \eqref{eqn:f_def}, with $g$ defined in \eqref{eqn:g_def}. Then
  \[
  \Var(\mathcal F)
  \;=\;
  \sup_{\phi:X\to[-1,1]^k}\;
      \mathbb E_{\pi \sim \Unif(S_m)}
      \Bigl[\,
        \ip{\phi,f_\pi-U_k}_{D}
      \Bigr]^{2}~~\le~~ \frac{1}{m!}  + e^2\,2^{-m} ~.
    \]
\end{theorem}

\subsection{Proof of \thmref{thm:main_mult}}

The proof of the theorem relies on \lemref{lem:parseval} (in \secref{sec:parseval}), which gives a generalization of Parseval's
inequality for functions with multiple outputs that are
$\epsilon$-approximately-orthonormal. 
In particular, the set of functions we will use is
\[
 \left\{ h_\pi
= \tfrac{k}{\sqrt{2}} (f_\pi
- U_k) ~:~ \pi \in S_m \right\}
\]
There are $m!$ such functions and we will see that they are
$\epsilon$-orthonormal with $\epsilon \le e^2\,2^{-m}$.

For every $x$,
\[
h_\pi(x) =  \tfrac{1}{\sqrt{2}} (e_{g(x_\pi)} -
e_{g(x_\pi)+5~\bmod~10})  ~.
\]
Therefore, 
$\inner{h_\pi(x),h_\pi(x)} = 1$ which yields $\inner{h_\pi,h_\pi}_D =
1$ as well. It is left to show that on average, for $\pi \neq
\sigma$ the value of $(\inner{h_\pi,h_\sigma}_D)^2$ is exponentially small.

\paragraph{Representing $h_\pi$ using Fourier characters:}
Let us first write the ten additive characters of the group
$\mathbb{Z}_{10}$ as
\[
\chi_\omega(z) = e^{2\pi i \omega z / k}, \quad \omega \in \Sigma
\]
Also, for every $\omega \in \Sigma$ define $v_\omega \in \mathbb{C}^{10}$
to be the vector whose $t$-th coordinate is $\chi_\omega(-t)$.
For $\omega,\eta \in \Sigma$, we have
\[
\frac{1}{k} \sum_{t=0}^9 \chi_\omega(t) \chi_\eta(-t) =
\delta_{\omega,\eta} ~.
\]
By a direct calculation one can verify that for every digit $z \in \Sigma$
\[
e_z - e_{z+5 \bmod 10} = \frac{2}{k} \sum_{\omega \in \{1,3,5,7,9\}} 
\chi_\omega(z) v_\omega
\]
It follows that
\[
h_\pi(x) = \frac{1}{\sqrt{2}} (e_{g(x_\pi)} - e_{g(x_\pi)+5~\bmod~10})
= \frac{\sqrt{2}}{k}
\sum_{\omega~\textrm{odd}} \chi_\omega(g(x_\pi))\,v_\omega
\]

\paragraph{Inner products become scalar characters:} 
Because the $v_\omega$ are orthogonal, with $\|v_\omega\|^2=k$,
\[
  \inner{h_\pi,h_\sigma}_D = \E_x \inner{h_\pi(x) , h_\sigma(x)} =
  \frac{2}{k^2}
  \E_x \sum_{\omega~\textrm{odd}} \chi_\omega(g(x_\pi))\,
  \overline{\chi_\omega(g(x_\sigma)) } \, k =  \frac{2}{k}
  \sum_{\omega~\textrm{odd}} \underbrace{\E_x \chi_\omega(g(x_\pi
    )-g(x_\sigma))}_{:= \Gamma_\omega(\pi,\sigma)}
\]
The next step is to bound $\Gamma_\omega(\pi,\sigma)$.

\paragraph{Separating the contribution of every pair of digits:}
For two different permutations $\pi,\sigma$, we have
\[
g(x_\pi)-g(x_\sigma) \bmod~10 = \sum_{i=0}^{m-1} 
(x_{\pi(i)}-x_{\sigma(i)}) x_{2m-1-i} ~\bmod~10
\]
Let
\[
t(\pi,\sigma) := |\{i : \pi(i) \neq \sigma(i)\}|
\]
be the size of the difference between the two permutations, and since
$\pi \neq \sigma$ we have $t(\pi,\sigma) \ge 2$. Because the $m$
digit-pairs are independent
\begin{align*}
\Gamma_\omega(\pi,\sigma)  &= \E_x \exp\left(2\pi j \tfrac{\omega}{k} \sum_{i : \pi(i) \neq \sigma(i)} (x_{\pi(i)}-x_{\sigma(i)}) 
x_{2m-1-i} \right) \\
&= \E_x \prod_{i : \pi(i) \neq \sigma(i)} \exp\left(2\pi j \tfrac{\omega}{k} (x_{\pi(i)}-x_{\sigma(i)}) 
x_{2m-1-i} \right) 
\end{align*}
Since the distribution over $x$ is symmetric, let's assume w.l.o.g.
that $\pi$ is the unit permutation. Also, we can ignore all
coordinates for which $\pi(i) = \sigma(i)$ as they aren't used in the
expression. Therefore, we can apply \lemref{lem:expectation_bound} in
\secref{app:multiTechnical} to get that:
\[
|\Gamma_\omega(\pi,\sigma)| \le 2^{-t(\pi,\sigma)}
\]

We have therefore shown that $\inner{h_\pi,h_\pi}_D = 1$ while for
$\pi \neq \sigma$ \[
  \inner{h_\pi,h_\sigma}_D ~\le~ \frac{2}{k}\,5 \cdot
  2^{-t(\pi,\sigma)}
  = 2^{-t(\pi,\sigma)}
\]
It follows that
\[
  \E_{\pi,\sigma} (\inner{h_\pi,h_\pi}_D - \delta_{\pi,\sigma})^2 ~\le~
  \E_{\pi,\sigma} \delta_{\pi,\sigma} \, (2^{-t(\pi,\sigma)})^2
\]
If two permutations differ, then $t(\pi,\sigma) \in \{2,...,m\}$.
Therefore, we can further write
\[
    \E_{\pi,\sigma} (\inner{h_\pi,h_\pi}_D - \delta_{\pi,\sigma})^2 \le
    \sum_{d=2}^m \prob[t(\pi,\sigma)=d]\,2^{-2d}
    = \sum_{d=2}^m \prob[t(\pi,\sigma)=d]\,4^{-d}
\]
\lemref{lem:t_bound} in \secref{app:multiTechnical} shows that for $d \in
\{2,...,m-1\}$
we have
that $\prob[t(\pi,\sigma)=d] < 1/(m-d)!$. Therefore,
\begin{align*}
  \E_{\pi,\sigma} (\inner{h_\pi,h_\pi}_D - \delta_{\pi,\sigma})^2
  &\le
    \sum_{d=2}^{m-1} \frac{4^{-d}}{(m-d)!} + 4^{-m} ~\le~
    e^4\,4^{-m}
\end{align*}
where in the last step we have used \lemref{lem:sum_bound}.

The proof of the theorem now follows from \lemref{lem:parseval}, where
we set $\epsilon = e^2\,2^{-m}$.

\subsection{Generalization of Parseval} \label{sec:parseval}

\begin{lemma}[Approximate Parseval for functions with multiple
  outputs] \label{lem:parseval}
Let $h_1,\ldots,h_n$ be a sequence of functions from a finite domain
$X$ to $\reals^k$. Define the Hilbert space $L^2(X,\reals^k)$ with the
inner product
\[
\inner{f,g}_H = \E_{x} \inner{f(x), g(x)}
\]
where the expectation is over a uniform $x$ from $X$ and $\inner{f(x),
  g(x)} = \sum_{j=1}^k f_j(x)g_j(x)$. Similarly, we
use $\|f\|_H^2 = \inner{f,f}$.
Let $E \in \reals^{n,n}$ be such that
\[
\inner{h_i,h_j}_H - \delta_{i,j} = E_{i,j}
\]
and denote
\[
\epsilon = \sqrt{ \E_i \E_j E_{i,j}^2} ~,
\]
where the expectations are over a uniform distribution over $[n]$.
Then,
\[
\E_i (\inner{h_i,\phi}_H)^2 \le \left(\frac{1}{n} + \epsilon\right)\,\|\phi\|_H^2
\]
where the expectation is over a uniform $i$ from $[n]$.
Or, in other words
\[
  \E_i \left(\E_x \inner{h_i(x),\phi(x)}\right)^2 \le
  \left(\frac{1}{n}
  + \epsilon\right)\,
E_x \inner{\phi(x),\phi(x)}
\]
\end{lemma}
\begin{proof}
  Denote $|X|=m$ and $d = mk$. 
We can think of each $f : X \to \reals^k$ as a vector in $\reals^d$ by listing for
every $x \in X$ (there are $m$ such $x$'s) the $k$ values of $f(x)$.
So, from now on, let's think of $h_1,\ldots,h_n$ and on $\phi$ as
vectors in $\reals^d$. For two functions $f,g$ both from $X \to
\reals^k$, we have
\[
  \inner{f,g}_H = \E_x \inner{f(x),g(x)} =
  \frac{1}{m}\,\inner{f,g}
  \]
where in the right-most term we refer to $f$ and $g$ as
$d$-dimensional vectors.

Let $H \in \reals^{n,d}$ be a matrix whose $i$-th row is the vector $h_i$. Let $\alpha
= H \phi$. We want to bound $\|\alpha\|_2^2$. We have
\[
\|\alpha\|_2^2 = \phi^\top H^\top H \phi
\]
This is bounded by $\lambda_{\max}(H^\top H) \|\phi\|^2$, where
$\lambda_{\max}$ is the maximal eigenvalue of $H^\top H$. Since the
maximal eigenvalue of $H^\top H$ is the same as that of the matrix $A
= H H^\top$, we shall bound the latter one by the following:
\[
  \lambda_{\max}(A) = \max_{v : \|v\|_2\le 1} v^\top A v =
  \max_{v : \|v\|_2\le 1} v^\top m(I + E) v \le
  m \max_{v : \|v\|_2\le 1} \left[ \|v\|_2^2 + \sum_{i=1}^n \sum_{j=1}^n v_i v_j E_{i,j}\right]
\]
By Cauchy-Schwartz
\[
\sum_{i=1}^n \sum_{j=1}^n v_i v_jE_{i,j} \le \sqrt{\sum_{i,j} v_i^2
  v_j^2}\,\sqrt{\sum_{i,j}  E_{i,j}^2} =
\sqrt{\sum_{i} v_i^2 \sum_j
  v_j^2}\,\sqrt{ n^2 \E_i \E_j
  E_{i,j}^2} = \|v\|_2^2\,n\,\epsilon
\]
All in all,
\[
\lambda_{\max}(A)  \le m(1 + n\,\epsilon)
\]
and so
\[
\|\alpha\|_2^2 \le \|\phi\|_2^2\,m\,(1+ n\,\epsilon) ~.
\]
Finally,
\begin{align*}
\E_i(\inner{h_i,\phi}_H)^2 &= \frac{1}{n} \sum_i
\left(\frac{\inner{h_i,\phi}}{m}\right)^2
                             =  \frac{1}{n} \, \frac{1}{m^2} \, \sum_i
                             (\inner{h_i,\phi})^2
=  \frac{1}{n} \, \frac{1}{m^2} \, \|\alpha\|_2^2
  \\
  &\le
\frac{1}{n} \, \frac{1}{m^2} \, 
\|\phi\|_2^2\, m\, (1+ n\,\epsilon)
    = \frac{1}{n} \, \|\phi\|_H^2(1 + n\,\epsilon)
    =  \left( \frac{1}{n} + \epsilon \right) ~\|\phi\|_H^2 ~,
\end{align*}
which concludes our proof.
\end{proof}

\subsection{Technical Lemmas for the Multiplication Hardness Proof} \label{app:multiTechnical}

\begin{lemma} \label{lem:t_bound}
  For two permutations, $\pi,\sigma \in S_m$, let
  $t(\pi,\sigma) = |\{i : \pi(i) \neq \sigma(i)\}|$. Then, for $2 \le d \le
  m-1$,
  \[
    \prob[t(\pi,\sigma)=d] < \frac{1}{(m-d)!}
  \]
  where the probability is over i.i.d. uniformly choice of $\pi$ and
  $\sigma$ from $S_m$.
\end{lemma}
\begin{proof}
  Clearly,
\[
\prob(t(\pi,\sigma)=d) = \frac{|\{(\pi,\sigma) \in S_m \times S_m :
  t(\pi,\sigma)=d\}|}{|S_m \times S_m|}
\]
For the numerator: Pick $\sigma$ first. There are $m!$ possibilities. Then pick the $m-d$
coordinates on which $\pi$ coincides with $\sigma$. There are ${m
  \choose (m-d)}$ options, which also equals to ${m \choose d}$.
Finally, given the $d$ ``mismatch'' positions, we need to pick $\pi$,
but there are at most $d!$ options for permutations over $d$ elements,
and one of them is $\sigma$, so the number of options is bounded by
$d!$. All in all,
\[
  |\{(\pi,\sigma) : t(\pi,\sigma)=d\}| < m!\, {m \choose d}\, d!
  = m!\, \frac{m!}{(m-d)!\,d!}\, d! = \frac{(m!)^2}{(m-d)!}
\]
Since $|S_m \times S_m| = (m!)^2$, we conclude the proof.
\end{proof}

\begin{lemma} \label{lem:sum_bound}
 For $d < m$,
  \[
    \sum_{d=2}^{m-1} \frac{4^{-d}}{(m-d)!} ~\le~ (e^4-1)\,4^{-m} ~.
  \]
\end{lemma}
\begin{proof}
Let $S$ denote the left-hand side. Let $k = m-d$. Then, as $d$ goes
from $2$ to $m-1$, $k$ goes from $1$ to $m-2$. The sum can be
rewritten as
\begin{align*}
  S & = \sum_{k=1}^{m-2} \frac{1}{k!} \cdot 4^{-m+k} \\
    &= 4^{-m}\, \sum_{k=1}^{m-2} \frac{4^k}{k!} \\
    &\le 4^{-m}\, \sum_{k=1}^{\infty} \frac{4^k}{k!} \\
  &= 4^{-m}\, (e^4-1)
\end{align*}
where the last equality is due to the Maclaurin series of $e^x$.
\end{proof}

\begin{lemma} \label{lem:expectation_bound}
Let \(k = 10\) and denote \([k] \coloneqq \{0,1,2,\dots ,9\}\).
Fix \(\omega \in \{1,3,5,7,9\}\) and an integer \(m\).
Let \(\sigma\) be a permutation of \([m]\) such that \(\sigma(i) \neq i\) for every \(i\).
Then
\[
\left|\,
\mathbb{E}_{a,b}\,
\exp\!\left(2\pi j\,\tfrac{\omega}{k}\,\sum_{i=0}^{m-1}(a_i - a_{\sigma(i)})\,b_i\right)
\,\right|
\;\le\;
2^{-m},
\]
where the expectation is over independent random vectors
\(a,b \in [k]^m\), each drawn uniformly from \([k]^m\).
\end{lemma}
\begin{proof}
For a fixed vector \(a\in[k]^m\) let
\[
I(a) \;=\;\bigl\{\,i\in[m] : a_i = a_{\sigma(i)}\,\bigr\},
\qquad
T(a) \;=\;[m] \setminus I(a).
\]
Define
\[
\alpha_i(b)
\;=\;
\exp\!\Bigl(2\pi j\,\tfrac{\omega}{k}\,(a_i-a_{\sigma(i)})\,b_i\Bigr),
\qquad i\in[m].
\]
so we need to bound $\left| \E_a \E_b \prod_i \alpha_i(b) \right|$. 
Because the coordinates of \(b\) are independent,
\[
\mathbb{E}_{a,b}\!\Bigl[\prod_{i=1}^{m}\alpha_i(b)\Bigr]
\;=\;
\mathbb{E}_{a}\!\Bigl[\;
\mathbb{E}_{b}\!\bigl[\prod_{i\in I(a)}\alpha_i(b)\bigr]
\;
\mathbb{E}_{b}\!\bigl[\prod_{i\in T(a)}\alpha_i(b)\bigr]
\Bigr].
\]
If \(i\in I(a)\) then \(a_i=a_{\sigma(i)}\) and hence \(\alpha_i(b)=1\).
If \(i\in T(a)\) the inner expectation vanishes because
\(\mathbb{E}_{b_i}\,\alpha_i(b)=0\) (the map
\(b_i\mapsto \alpha_i(b)\) has mean zero on the uniform distribution
over \([k]\)).
Consequently
\[
\mathbb{E}_{a,b}\!\Bigl[\prod_{i=1}^{m}\alpha_i(b)\Bigr]
\;=\;
\Pr_{a}\!\bigl[T(a)=\varnothing\bigr]
\;=\;
\Pr_{a}\!\bigl[a_i=a_{\sigma(i)}\ \forall\,i\bigr].
\]

\smallskip
\noindent\textbf{Cycle decomposition.}
Write the derangement \(\sigma\) as the product of \(c=c(\sigma)\) disjoint cycles
\[
\sigma \;=\;
(i_{1,1}\,i_{1,2}\,\dots\,i_{1,\ell_1})
\cdots
(i_{c,1}\,i_{c,2}\,\dots\,i_{c,\ell_c}),
\]
with \(\ell_r\ge 2\) and \(\sum_{r=1}^{c}\ell_r=m\).
Inside a single cycle all coordinates must coincide:
\(a_{i_{r,1}}=\cdots=a_{i_{r,\ell_r}}\).
For that cycle one may choose the common value freely (10 choices) and
each of the remaining \(\ell_r-1\) coordinates must match it,
each with probability \(1/10\).  Thus
\[
\Pr\bigl[a\text{ is constant on cycle }r\bigr]=10^{-(\ell_r-1)}.
\]
Independence across cycles gives
\[
\Pr_{a}\!\bigl[a_i=a_{\sigma(i)}\ \forall\,i\bigr]
\;=\;
\prod_{r=1}^{c}10^{-(\ell_r-1)}
\;=\;
10^{-\sum_{r=1}^{c}(\ell_r-1)}
\;=\;
10^{-(m-c)}.
\]

\smallskip
\noindent\textbf{Exponential decay.}
Every cycle has length at least \(2\), hence \(c\le\lfloor m/2\rfloor\).
Therefore
\[
10^{-(m-c)}
\;\le\;
10^{-(m-\lfloor m/2\rfloor)}
\;=\;
10^{-\lceil m/2\rceil}
=
\bigl(10^{-1/2}\bigr)^{m}
\le
\bigl(2^{-1}\bigr)^{m}.
\]
Combining all steps yields the claimed bound
\(
\bigl|\mathbb{E}_{a,b}\exp(\dots)\bigr|\le 2^{-m}.
\)
\end{proof}

\section{The Shortest Path Example} \label{sec:shortest}

The following lemma characterizes the optimal solution of the shortest
path problem and relate it to parity.
\begin{lemma}[Parity\label{lem:parity}]
For every $\pi\in S_n$ and every $x\in\{\pm1\}^n$ let $f_\pi(x)$ be
$1$ if the shortest path from $s$ to $t$ goes through $a_0$ and
$f_\pi(x)=-1$ if the shortest path goes through $b_0$. Then,
$f_\pi(x)$ can be rewritten as
\[
  f_\pi(x)\;=\;\prod_{\substack{1\le j\le n\\ j\text{ odd}}}x_{\pi(j)}.
\]
That is, $f_\pi(x)=+1$ iff an even number of the bits read in the
\emph{odd} positions of the order $\pi$ are $-1$.
Furthermore, for the rest of the nodes in the shortest path, the rule
``pick an edge with a zero weight'' leads to an optimal shortest path.
\end{lemma}
\begin{proof}  
Let
\[
  T(x,\pi)=\bigl|\{\,j\le n: j\text{ odd and }x_{\pi(j)}=-1\,\}\bigr|
\]
be the number of minus--bits encountered in odd layers. 
In every even layer the path can proceed straight at zero cost, so row switches are relevant only in odd layers.  In an odd layer the cheaper option is to \emph{stay} if $x_{\pi(j)}=+1$ and to \emph{switch} if $x_{\pi(j)}=-1$.  
\begin{itemize}
\item If $T(x,\pi)$ is even, let us start a path on the top row $(s,a_0)$, and from there on, for every even layer we simply continue on the same layer and for every odd layer we choose the edge with a weight of $0$. This means that we switch layers an even number of times, so we end up at the node $a_n$, from which we move to node $t$. The total cost of this path is $0$, and this must be optimal since there are no negative weights. In contrast, if we would start with $(s,b_0)$ there is at least one time we need to pay a cost of $1$ for moving between the layers. Therefore, the shortest path begins with $(s,a_0)$ and $f_\pi(x)=+1$.
  \item If $T(x,\pi)$ is odd, the situation is reversed and $f_\pi(x)=-1$.
\end{itemize}
Finally, $(-1)^{T(x,\pi)}=\prod_{j\text{ odd}}x_{\pi(j)}$, which proves the claim.
\end{proof}

\section{Failure of Tree-of-Thoughts}

Below we show that both variants of the ToT framework, ToT-BFS and
ToT-DFS are likely to fail on the boosted version of
Example~\ref{example:drift}.

Recall the boosted instance of Example~\ref{example:drift} as in the
main text: a random matrix $x\!\in\!\{\pm1\}^{2n\times n}$ with
column-wise permutations $\pi_j$, and latent states
\[
   v_{2,j}=\prod_{i=1}^n x_{\pi_j(i),j},\qquad
   v_{i,j}=v_{i-1,j}\,x_{\pi_j(i),j}\;(i \ge 3),
\]
with target output $v_{n-2}\in\{\pm1\}^n$.

Since the family of parity functions is orthonormal, using the same
arguments as in \secref{sec:multiplication}, without performing an
exponentially large number of SGD iterations, for any $t < n/2$, the
distribution of $v_t$ looks to the learner as uniform. We rely on this
fact for deriving the failure of ToT.

\subsection{Failure of ToT-BFS} \label{sec:failure_BFS_TOT}
In ToT-BFS, the search procedure can maintain at most $B$ options for
each level. It selects which of the options to keep by expanding the
previous layer and then using a value network to score them. However,
as argued before, for $t < n/2$, all options looks to the value
network as uniform. Hence, the chance of picking the right options is
exponentially small, leading to the failure of ToT-BFS.

\subsection{Failure of ToT-DFS} \label{sec:failure_DFS_TOT}

In ToT-DFS, the learner creates a full path until it either succeed or
the value function tells it that a backtrack is required. As argued
before, the probability to succeed is exponentially small. When
backtrack is issued, it is not learnt but rather, we go one step
backward. This means that we basically search randomly in an
exponentially large space, which would lead to success only after an
exponentially high number of search operations.

\section{Technical Lemmas} \label{sec:technical}

\begin{lemma} \label{lem:geinq}
  For $\gamma \in (0,1)$ and $\epsilon \in (0,\gamma)$ we have that 
\[
  \frac{\gamma(1-\epsilon)}{\gamma(1-\epsilon)+\epsilon} ~\ge~
(1-\tfrac{\epsilon}{\gamma}) ~.
\]
\end{lemma}
\begin{proof}
  Since $\gamma \le 1$ we have
  \[
0 \ge \epsilon^2(1-\tfrac{1}{\gamma}) = \epsilon - \epsilon(1-\epsilon)-\frac{\epsilon^2}{\gamma}
\]
Adding $\gamma(1-\epsilon)$ to both sides we obtain
\[
\gamma(1-\epsilon) ~\ge~ \gamma(1-\epsilon) + \epsilon -
\epsilon(1-\epsilon)-\frac{\epsilon^2}{\gamma} ~=~ (\gamma(1-\epsilon)+\epsilon)(1-\tfrac{\epsilon}{\gamma})
\]
Dividing both sides by $\gamma(1-\epsilon)+\epsilon$ we conclude our proof.
\end{proof}

\bibliographystyle{plainnat}
\bibliography{SI}

\end{document}